\documentclass[10pt,journal,compsoc]{IEEEtran}
\ifCLASSOPTIONcompsoc
  \usepackage[nocompress]{cite}
\else
  \usepackage{cite}
\fi
\ifCLASSINFOpdf
\else
\fi
\usepackage{times}
\usepackage{epsfig}
\usepackage{graphicx}
\usepackage{amsmath}
\usepackage{amssymb}
\usepackage[numbers,sort]{natbib}
\usepackage[switch]{lineno}
\usepackage{tabulary}
\usepackage{amsmath}
\usepackage{tabularx}
\usepackage{booktabs}
\usepackage{adjustbox}
\usepackage{multirow}
\usepackage[flushleft]{threeparttable}
\usepackage{makecell}
\usepackage{tabulary}
\usepackage{amsmath}
\usepackage{amsfonts} 
\usepackage{pifont}
\usepackage{kotex}
\newcommand{\xmark}{\text{\ding{55}}}
\newcommand{\cmark}{\text{\ding{51}}}
\newcommand\rv[1]{{\color{black}#1}}
\newcommand\jmrv[1]{{\color{black}#1}}
\newcommand\rvt[1]{{\color{black}#1}}

\usepackage{bbm}
\usepackage[dvipsnames]{xcolor}

\usepackage[colorlinks=true,linkcolor=black, citecolor=black]{hyperref}

\usepackage{caption}

\usepackage{url}
\begin{document}
\title{Anti-Adversarially Manipulated Attributions for Weakly Supervised Semantic Segmentation and Object Localization}

% \author{Michael~Shell,~\IEEEmembership{Member,~IEEE,}
%         John~Doe,~\IEEEmembership{Fellow,~OSA,}
%         and~Jane~Doe,~\IEEEmembership{Life~Fellow,~IEEE}% <-this % stops a space
\author{Jungbeom Lee, Eunji Kim, Jisoo Mok, and Sungroh Yoon,~\IEEEmembership{Senior Member,~IEEE}\IEEEcompsocitemizethanks{\IEEEcompsocthanksitem J. Lee, E. Kim, J. Mok and S. Yoon are with the Department of Electrical and Computer Engineering, Seoul National University, Seoul 08826, South Korea.\\
% E-mail: see http://www.michaelshell.org/contact.html
\vspace{-1em}
\IEEEcompsocthanksitem S. Yoon is also with Interdisciplinary Program in AI, AIIS, ASRI, INMC, and ISRC, Seoul National University, Seoul 08826, South Korea. \\
\vspace{-1em}
\IEEEcompsocthanksitem S. Yoon is the corresponding author (sryoon@snu.ac.kr).}}% <-this % stops an unwanted space}
% \thanks{Manuscript received April 19, 2005; revised August 26, 2015.}}


\markboth{IEEE TRANSACTIONS ON PATTERN ANALYSIS AND MACHINE INTELLIGENCE}%
{Shell \MakeLowercase{\textit{et al.}}: Bare Demo of IEEEtran.cls for Computer Society Journals}



\IEEEtitleabstractindextext{%
\begin{abstract}
Obtaining accurate pixel-level localization from class labels is a crucial process in weakly supervised semantic segmentation and object localization.
Attribution maps from a trained classifier are widely used to provide pixel-level localization, but their focus tends to be restricted to a small discriminative region of the target object. An AdvCAM is an attribution map of an image that is manipulated to increase the classification score produced by a classifier \rv{before the final softmax or sigmoid layer}. This manipulation is realized in an anti-adversarial manner, so that the original image is perturbed along pixel gradients in directions opposite to those used in an adversarial attack. 
This process enhances non-discriminative yet class-relevant features, which make an insufficient contribution to previous attribution maps, so that the resulting AdvCAM identifies more regions of the target object. 
In addition, we introduce a new regularization procedure that inhibits the incorrect attribution of regions unrelated to the target object and the excessive concentration of attributions on a small region of the target object.
Our method achieves a new state-of-the-art performance in weakly and semi-supervised semantic segmentation, on both the \rv{PASCAL VOC 2012} and \rv{MS COCO 2014} datasets.
In weakly supervised object localization, 
it \rvt{achieves} a new state-of-the-art performance on the CUB-200-2011 and ImageNet-1K datasets.

\end{abstract}

}

\maketitle

\IEEEdisplaynontitleabstractindextext

% \fi
\IEEEpeerreviewmaketitle

\IEEEraisesectionheading{\section{Introduction}\label{sec:introduction}}


% \IEEEPARstart{}{bject} recognition and scene understanding is one of the most important tasks in the computer vision society.
\newcommand{\RNum}[1]{\lowercase\expandafter{\romannumeral #1\relax}}

\IEEEPARstart{U}{nderstanding} the semantics of an image and recognizing objects in it are vital processes in computer vision systems. These tasks involve semantic segmentation, in which a semantic label is allocated to each pixel of an image, and object localization, otherwise known as single-object detection, which locates a target object in \jmrv{the} form of a bounding box. 
Although deep neural networks (DNNs) have facilitated tremendous progress in both tasks~\cite{chen2017deeplab, huang2019ccnet, chen2018encoder, ren2015faster, redmon2016you, lin2017focal}, it has come at the cost of annotating thousands of training images with explicit localization cues. In particular, for semantic segmentation, pixel-level annotation of images containing an average of 2.8 objects takes about 4 minutes per image~\cite{bearman2016s}; and a single large (2048$\times$1024) image depicting a complicated scene requires more than 90 minutes for pixel-level annotation~\cite{cordts2016cityscapes}.

The need for such expensive annotations is sidestepped by weakly supervised learning, in which a DNN is trained on images with \jmrv{some form of abbreviated annotation \rvt{that} is cheaper than explicit localization cues.} 
Weakly supervised semantic segmentation methods can use scribbles~\cite{tang2018normalized}, points~\cite{bearman2016s}, bounding boxes~\cite{khoreva2017simple, song2019box, lee2021bbam}, or class labels~\cite{lee2019ficklenet, lee2022weakly, ahn2018learning, chang2020weakly, kim2022bridging} as annotations. 
The last of these are the cheapest and most popular option, largely because the images in many public datasets are already annotated with class labels~\cite{deng2009imagenet, everingham2010pascal}, and automated web searches can also provide images with class labels~\cite{lee2019frame, hong2017weakly, shen2018bootstrapping}.
Likewise, in weakly supervised object localization, class labels are a popular choice of annotation for localizing target objects with bounding boxes.
Weakly supervised semantic segmentation and object localization share the same goal, inasmuch as their aim is to generate informative localization cues that allow the regions occupied by a target object to be identified with class labels.

\begin{figure}[t]
\centering
\includegraphics[width=\linewidth]{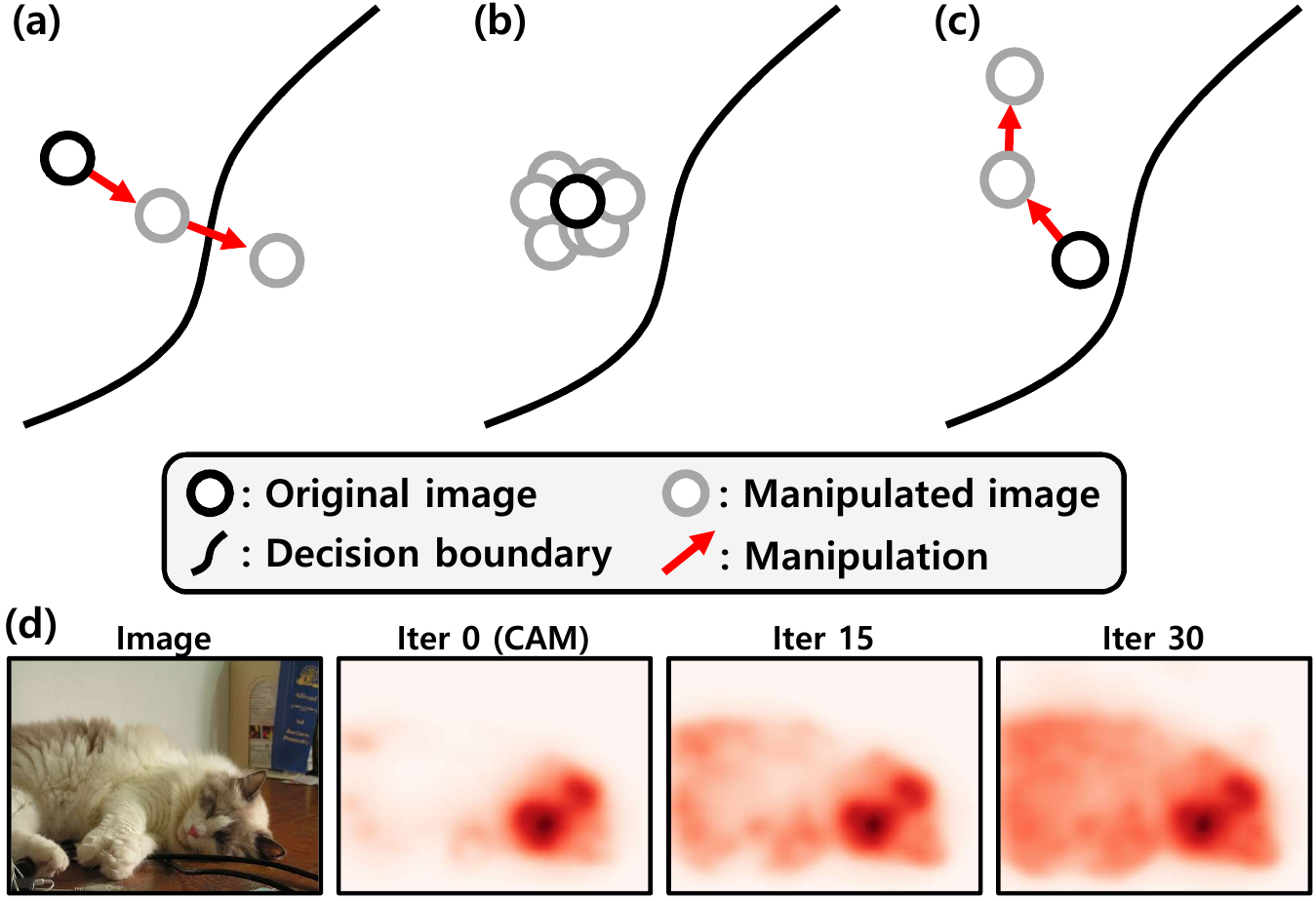}
\vspace{-2em}
\caption{\label{overview} Overview of image manipulation methods for weakly supervised semantic segmentation and object localization: (a) erasure~\cite{wei2017object, hou2018self, zhang2018adversarial}; (b) FickleNet~\cite{lee2019ficklenet}; and (c) AdvCAM. (d) Examples of successive \rv{localization maps} obtained from images iteratively manipulated by our method.}
\vspace{-0.5em}
\end{figure}

%\begin{figure}[!t]
%\centering
%\includegraphics[width=2.5in]{myfigure}
% where an .eps filename suffix will be assumed under latex, 
% and a .pdf suffix will be assumed for pdflatex; or what has been declared
% via \DeclareGraphicsExtensions.
%\caption{Simulation results for the network.}
%\label{fig_sim}
%\end{figure}
Most weakly supervised semantic segmentation and object localization methods depend on attribution maps obtained from a trained classifier, such as a Class Activation Map (CAM)~\cite{zhou2016learning} or a Grad-CAM~\cite{selvaraju2017grad}.
An attribution map identifies the important, or discriminative, regions of an image on which the classifier has concentrated.
But these regions tend to be relatively small, and most attribution maps do not identify the whole region occupied by \jmrv{the} target object.
Therefore, many researchers have tried to extend attributed regions to cover more of \jmrv{the} target object, by manipulating either the image~\cite{wei2017object, li2018tell, singh2017hide, babar2021look} or the feature map~\cite{lee2019ficklenet, zhang2018adversarial, hou2018self, choe2020attention, ki2020sample}. 

%%%%%%%%%%%%%%%%%%%%%%%%%%%
One popular method of manipulation is erasure, which iteratively removes the discriminative regions which have already been identified, forcing the classifier to find new regions of the target object~\cite{wei2017object, li2018tell, babar2021look, singh2017hide, hou2018self, zhang2018adversarial, mai2020erasing, ki2020sample}.
Erasure is effective, but if the image from which the discriminative region has been erased crosses the decision boundary, as shown in Fig.~\ref{overview}(a), an erroneous attribution map may be generated.
\rvt{An alternative manipulation method} is \rvt{a} stochastic perturbation, shown in Fig.~\ref{overview}(b): FickleNet~\cite{lee2019ficklenet} produces diverse attribution maps from an image by applying random dropout to the feature maps obtained by a \rvt{DNN and} then aggregates them into a unified map. 

We propose a new manipulation method for extending the attributed regions of a target object.
Our method is a benign twist on the established technique of adversarial attack~\cite{goodfellow2014explaining, kurakin2016adversarial}, which operates by finding a small perturbation of an image that pushes it across the decision boundary to change the classification result.
Our method operates in a reversed, or `anti-adversarial' manner: 
it aims to find a perturbation that drives the manipulated image \textit{away from} the decision boundary, as shown in Fig.~\ref{overview}(c).
\rv{This manipulation is realized by `adversarial climbing', which perturbs an image along pixel gradients so as to increase the logit of the target class produced by the classifier before its \rvt{softmax or sigmoid} layer.
From here on, we refer to the value of this logit as the classification score of the target class.}
As a result of iterative adversarial climbing, the CAM of the manipulated image gradually identifies more regions of the target object, as shown in Fig.~\ref{overview}(d). 
The attribution maps obtained from images that have been iteratively manipulated in this way can be used as localization cues for weakly supervised semantic segmentation and object localization.

While ascending the gradient ensures that the classification score increases, excessive ascent may cause problems: (\RNum{1}) irrelevant \rv{regions}, such as parts of the background or regions of other objects, can also be activated, or (\RNum{2}) the attribution scores of some parts of the target object can \rv{get unreasonably large}.
We address these problems by introducing regularization terms that suppress the scores of classes other than the target class and limit the attribution scores of regions that already have high scores.

Existing methods that aim to expand the identified region of a target object require additional modules~\cite{wang2020self, lee2019ficklenet, lu2020geometry, mai2020erasing} or a different training technique~\cite{chang2020weakly, zhangsplitting, ki2020sample, zhang2020inter}.
Our method is a post-hoc analysis of the trained classifier, and therefore can be used to improve the performance of existing methods without modification. It achieves a new state-of-the-art performance on the PASCAL VOC 2012~\cite{everingham2010pascal} and \rv{MS COCO 2014}~\cite{lin2014microsoft} datasets in both weakly and semi-supervised semantic segmentation. In weakly supervised object localization, our method \rvt{again} achieves a new state-of-the-art performance on the CUB-200-2011~\cite{wah2011caltech} and ImageNet-1K~\cite{deng2009imagenet} datasets.

This paper is an extended version of our previous \rv{publication}~\cite{lee2021anti}.
In this paper, (\RNum{1}) we include more detailed explanations of our method\rv{;} (\RNum{2}) we propose a new technique, which seamlessly integrates a salient object detector into adversarial climbing for weakly supervised semantic segmentation\rv{;}
(\RNum{3}) we present additional experimental results on weakly supervised semantic segmentation, including tests \rv{on more backbones and an additional dataset (\rv{MS COCO 2014}~\cite{lin2014microsoft})}\rv{;} (\RNum{4}) we also test our algorithm on weakly supervised object localization with the ImageNet-1K~\cite{deng2009imagenet} and CUB-200-2011~\cite{wah2011caltech} datasets, with the intention of demonstrating the wide applicability of adversarial climbing\rv{; and (\RNum{5}) we analyze our method from various viewpoints and provide deeper insights into the properties of AdvCAM}.

The main contributions of this paper can be summarized as follows:
\begin{itemize}
\vspace{-3pt}
	\item[$\bullet$] We propose AdvCAM, an attribution map of an image that is manipulated to increase the classification score, allowing it to identify more of the \rv{region} occupied by an object.
% 	\vspace{-5pt}
    \rv{\item[$\bullet$] We demonstrate the generality of our method by showing that 1) it can be seamlessly integrated with saliency supervision; 2) it can be applied to multiple tasks, namely weakly and semi-supervised semantic segmentation and weakly supervised object localization; and 3) it improves the performance of several existing methods for weakly supervised semantic segmentation and object localization, without modification or re-training of their networks.}
    
    \rv{\item[$\bullet$] We show that our method produces a significantly better performance on the PASCAL VOC 2012 and MS COCO 2014 datasets than existing methods, in both weakly and semi-supervised semantic segmentation. We also achieve new state-of-the-art results on the CUB-200-2011 and ImageNet-1K datasets in weakly supervised object localization.}
    
    \rv{\item[$\bullet$] We analyze our method from various viewpoints, providing deeper insights into the properties of AdvCAM.}
\end{itemize}

\section{Related Work}

\subsection{Weakly Supervised Semantic Segmentation}\label{re_weak}
\rv{The common pipeline for weakly supervised semantic segmentation \rvt{consists of three main processes}: 1) obtaining initial seed regions, \textit{e.g.,} by using a CAM~\cite{zhou2016learning}, 2) producing pseudo ground truth masks by refining the initial seed, and 3) training a segmentation network with the resulting pseudo ground truth.}

\textbf{Obtaining a High-Quality Seed:}
Several methods have been proposed to improve the quality of the initial seed regions obtained from classifiers. 
Wei \textit{et al.}~\cite{wei2017object} obtain new attribution maps using images from which \rv{the previously identified discriminative regions} have been erased.
Other researchers have embedded this erasure technique into \rv{their} training \rv{schemes}~\cite{li2018tell, hou2018self}.
Wang \textit{et al.}~\cite{wang2020self} use equivariance regularization during the training of their classifier, so that the attribution maps obtained from differently transformed images are equivariant to those transformations.
Chang \textit{et al.}~\cite{chang2020weakly} improve feature learning by using latent semantic classes that are sub-categories of annotated parent classes, which can be pseudo-labeled by clustering image features.
Zhang \textit{et al.}~\cite{zhangsplitting} produce two different attribution maps from different classifiers and aggregate them into a single map.
Fan \textit{et al.}~\cite{fan2018cian} and Sun \textit{et al.}~\cite{sun2020mining} capture information shared among several images by considering cross-image semantic similarities and differences. 
\rv{Zhang \textit{et al}.~\cite{zhang2020causal} analyze \jmrv{the} co-occurrence context problem in multi-label classification and propose context adjustment (CONTA) to remove the confounding bias, resulting in a CAM seed free \jmrv{of} spurious correlations.}
Wei \textit{et al.}~\cite{wei2018revisiting} and Lee \textit{et al.}~\cite{lee2018robust} consider the target object in several contexts by combining multiple attribution maps obtained from differently dilated convolutions or from different layers of \rvt{a} DNN.

\textbf{Growing the Object Region:}
Some researchers expand an initial seed using a method analogous to region growing, in which they examine the neighborhood of each pixel.
They first use a CAM~\cite{zhou2016learning} to identify seed \rv{regions} \rvt{that} can confidently be associated with the target object. 
Semantic labels are then propagated from those confidently identified regions to ambiguous regions of the CAM which initially had low confidence scores.
SEC~\cite{kolesnikov2016seed} and DSRG~\cite{huang2018weakly} allocate pseudo labels to those ambiguous regions using \rv{a conditional random field (CRF)}~\cite{krahenbuhl2011efficient} during the training of the segmentation network.
PSA~\cite{ahn2018learning} and IRN~\cite{ahn2019weakly} train a DNN to capture the relationship between pixels and then propagate the semantic labels of confidently identified regions to semantically similar \rv{regions} by a random walk.
\rv{BES}~\cite{chenweakly} synthesizes a pseudo boundary from a CAM~\cite{zhou2016learning} and then uses a similar semantic propagation process to that of PSA~\cite{ahn2018learning}.

\subsection{Semi-Supervised Semantic Segmentation}\label{re_semi}
In semi-supervised learning, a segmentation network is trained using a small number of images with pixel-level annotations, together with a much larger number of images with weak or no annotations.
\jmrv{Cross-consistency training (CCT)~\cite{ouali2020semi} enforces \rv{the} invariance of predictions over a range of perturbations such as random noise and spatial dropout.}
Lai \textit{et al.}~\cite{Lai2021semi} enforce consistency among the features of the same object occurring in different contexts.
Luo \textit{et al.}~\cite{luosemi} introduce a network equipped with two separate branches, one of which is trained with strong labels and the other with weak labels.
Zou \textit{et al.}~\cite{Lai2021semi} design a pseudo-labeling process to calibrate the confidence score of pseudo labels for unlabeled data.
Souly \textit{et al.}~\cite{souly2017semi} use images synthesized by a generative adversarial network~\cite{goodfellow2014generative}, which improves feature learning.
Hung \textit{et al.}~\cite{hung2019adversarial} adopt an adversarial training scheme that increases the similarity of the distribution of the predicted segmentation maps to that of ground-truth maps.

\subsection{Weakly Supervised Object Localization}\label{re_wsol}
\rv{Weakly supervised object localization aims to predict the bounding box of a target object using class labels.} 
Most methods for weakly supervised object localization use \jmrv{a similar type of the CAM~\cite{zhou2016learning} \rvt{to} that used in weakly supervised semantic segmentation.}
HaS~\cite{singh2017hide} removes random rectangular patches \rv{from} a training image, forcing the classifier to \rv{examine} other regions of the target object.
ACoL~\cite{zhang2018adversarial} has two separate branches: \jmrv{one branch identifies the discriminative regions of an object and erases them based on features, and the other branch finds complementary regions from those erased features.}    
ADL~\cite{choe2020attention} and the technique introduced by Ki \textit{et al.}~\cite{ki2020sample} perform erasure realized by dropout during the training of a classifier.
Babar \textit{et al.}~\cite{babar2021look} combine the information from two intermediate images produced by regional dropout at complementary spatial locations.
CutMix~\cite{yun2019cutmix} is a data augmentation technique that combines two patches from different images and \jmrv{assigns} a new class label, which reflects the areas of the patches, to the resulting image.
Most methods of weakly supervised object localization share a single network for classification and detection, but GC-Net~\cite{lu2020geometry} uses a separate network for each task.

\section{Preliminaries}

\subsection{Adversarial Attack}\label{adv_attack_method}

An adversarial attack attempts to fool a DNN by presenting it with images that have been manipulated with intent to deceive.
Adversarial attack can be applied to classifiers~\cite{goodfellow2014explaining, moosavi2016deepfool}, semantic segmentation networks~\cite{arnab2018robustness}, or object detectors~\cite{xie2017adversarial}.
Not \jmrv{only} the predictions of a DNN, but \jmrv{also the} attribution maps can be altered by adversarial image manipulation~\cite{dombrowski2019explanations} or model parameter manipulation~\cite{heo2019fooling}.
These types of \rvt{attacks} try to make the DNN produce a spurious attribution map \rvt{that} identifies \jmrv{a} wrong location in the image, or a map \rvt{that} might have been obtained from a completely different image, without significantly changing the output of the DNN.

An adversarial attack on a classifier aims to find a small pixel-level perturbation that can change its \jmrv{decision.} 
% classifications.
In other words, given an input $x$ to the classifier, the adversarial attack aims to find the perturbation $n$ that \rvt{satisfies} $\texttt{NN}(x) \neq \texttt{NN}(x+n)$, where $\texttt{NN}(\mathord{\cdot})$ is the classification output \jmrv{from} the DNN.
A representative method~\cite{goodfellow2014explaining} of constructing $n$ for an attack starts by constructing the vector normal to the decision boundary of $\texttt{NN}(x)$, which can be realized by finding the gradients of $\texttt{NN}(x)$ with respect to $x$. 
A manipulated image $x'$ can then be obtained as follows:
\vspace{-0.2em}
\begin{linenomath}\begin{align}\label{adv_attack}
x' = x - \xi \nabla_x \texttt{NN}(x),
\vspace{-0.8em}
\end{align}\end{linenomath}
where $\xi$ determines the extent of the change to the image. This process can be understood as performing gradient descent on the image. 
PGD~\cite{kurakin2016adversarial}, which is a popular method of adversarial attack, performs the manipulation of Eq.~\ref{adv_attack} iteratively.

\subsection{Class Activation Map}\label{CAM_method}
A CAM~\cite{zhou2016learning} identifies the region of an image on which a classifier has concentrated.
It is computed from the class-specific contribution of each channel of the feature map to the classification score. A CAM is based on a convolutional neural network that has global average pooling (GAP) before its last classification layer. 
\jmrv{This process can be expressed as follows:}
\begin{linenomath}\begin{equation}\label{cam}
\texttt{CAM}(x) = \mathbf{w}^\intercal_c f(x),
\end{equation}\end{linenomath}
where $x$ is the image, $\mathbf{w}_c$ is the weight of the final classification layer for class $c$, and $f(x)$ is the feature map of $x$ prior to GAP.

A CAM bridges the gap between image-level and pixel-level annotation. However, the regions obtained by a CAM are usually much smaller than the full extent of the target object, since the small discriminative regions provide sufficient information for classification.

\section{Proposed Method}

\subsection{Adversarial Climbing}\label{Advcam_method}
AdvCAM is an attribution map obtained from an image manipulated using adversarial climbing, which perturbs the image in an anti-adversarial manner that is designed to increase the classification score of the image.
This is the reverse of an adversarial attack based on Eq.~\ref{adv_attack}, which manipulates the image to reduce the classification score.

Inspired by PGD~\cite{kurakin2016adversarial}, iterative adversarial climbing of an initial image $x^{0}$ can be performed using the following relation:
\vspace{-0.1em}
\begin{linenomath}\begin{equation}\label{sgd}
x^{t} = x^{t-1} + \xi \nabla_{x^{t-1}} y_c^{t-1},
\vspace{-0.2em}
\end{equation}\end{linenomath}
where $t$ ($1\leq t \leq T$) is the adversarial step index, $x^{t}$ is the manipulated image after step $t$, and $y_c^{t-1}$ is the classification logit of $x^{t-1}$ for class $c$, which is the output of the classifier before the final softmax or sigmoid layer.

This process enhances non-discriminative yet class-relevant features, which previously made insufficient contributions to the attribution map. 
Therefore, the attribution map obtained from an image manipulated by iterative adversarial climbing gradually identifies more regions of the target object.
More details of how adversarial climbing improves CAMs as intended are provided in Section~\ref{how_advcam}.
\rv{Noise can be expected to be introduced during late adversarial iterations, and this can be suppressed by producing the final localization map\footnote{\rv{Note that we will refer to an attribution map that has been subject to further processing as a localization map.}} $\mathcal{A}$ from an aggregation of the CAMs obtained from the manipulated images produced at each iteration $t$, as follows:}
\vspace{-0.3em}
\begin{linenomath}\begin{align}\label{aggregate}
\mathcal{A} = 
\frac{\sum_{t=0}^{T} \texttt{CAM}(x^{t})}{\max \sum_{t=0}^{T} \texttt{CAM}(x^{t})}.
\vspace{-0.5em}
\end{align}\end{linenomath}

\subsection{How Can Adversarial Climbing Improve CAMs?}\label{how_advcam}
When adversarial climbing increases $y_c$, it also increases the pixel values in the CAM, \jmrv{as can be} inferred from the relationship between a classification logit $y_c$ and a CAM (\textit{i.e.} $y_c = \text{GAP}(\texttt{CAM})$~\cite{zhang2018adversarial}).
Subsequently, we see from Eq.~\ref{cam} that an increase in the pixel values in the CAM will enhance some features.
\rv{If it is to produce better localization, adversarial climbing must meet \rvt{the} following conditions:}
(\RNum{1}) it enhances non-discriminative features, and (\RNum{2}) those features are class-relevant from a human point of view.
We analyze these two aspects of adversarial climbing in the following sections.

\subsubsection{How are non-discriminative features enhanced?} \label{nondisc_enahnce}
As the DNN's receptive field grows with an increasing number of layers, a change to one pixel in an input image \rvt{propagates} to many intermediate features.
\rv{This} propagation may affect both discriminative and non-discriminative features. 
\jmrv{Using the concept of strongly and weakly correlated features introduced by Tsipras \textit{et al.}~\cite{tsipras2018robustness} and Ilyas \textit{et al.}~\cite{ilyas2019adversarial}, we investigate how adversarial climbing can enhance non-discriminative features.}
\rv{Individually, each} weakly correlated feature may be of little importance to the corresponding class, but an accumulation of such features can greatly influence \rv{the classification result}.
It has been argued~\cite{tsipras2018robustness, ilyas2019adversarial} that an adversarial attack is made possible because a small change along the pixel gradient to an image changes many weakly correlated features to produce an erroneous classification.
Because adversarial climbing is the reverse of an adversarial attack, it can also be expected to \rv{significantly influence weakly correlated} (or non-discriminative) features.

\begin{figure}[t]
\centering
\includegraphics[width=\linewidth]{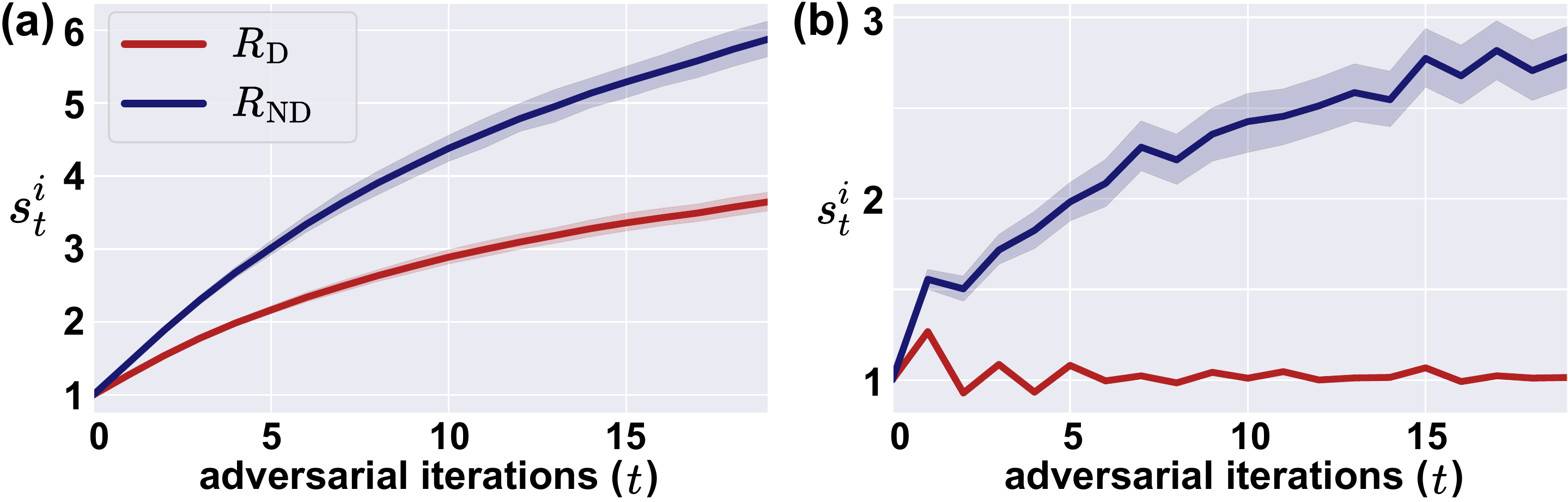}
\vspace{-2em}
\caption{\label{fig_amp} Distributions of the pixel amplification ratio $s^i_t$ for $i \in R_\text{D}$ and $i \in R_{\text{ND}}$, for 100 images, (a) without regularization and (b) with regularization.}
\vspace{-1em}
\end{figure}

We support this analysis empirically.
We define the discriminative region $R_\text{D} \!=\! \{i|\texttt{CAM}(x^{0})_i\!\geq\!0.5\}$ and the non-discriminative region $R_{\text{ND}} \!=\! \{i|0.1\!<\!\texttt{CAM}(x^{0})_i\!<\!0.5\}$\footnote{\rv{We set the lower bound to 0.1 because this value was found to exclude most of the background, which should not be considered in this analysis.}}, where $i$ is the location index. 
The pixel amplification ratio $s^i_t$ is $\texttt{CAM}(x^{t})_i/\texttt{CAM}(x^{0})_i$ at location $i$ and step $t$.
Fig.~\ref{fig_amp}(a) shows that adversarial climbing \rv{causes} both $s^{i \in {R_\text{D}}}_t$ and $s^{i \in {R_{\text{ND}}}}_t$ \rv{to} grow, but \rv{it also} enhances non-discriminative features more than discriminative ones, producing a descriptive CAM that identifies more regions of the target object.

\subsubsection{Are these enhanced features class-relevant?}
We now \rv{examine} whether the non-discriminative features identified by our technique are class-relevant from a human point of view.
\jmrv{When} considering a loss landscape with respect to an input, Moosavi \textit{et al.}~\cite{moosavi2019robustness} argued that a sharply curved loss landscape makes the input vulnerable to an adversarial attack.
Conversely, inputs that exist on a flat loss landscape are known~\cite{moosavi2019robustness, qin2019adversarial} to be robust against adversarial manipulations.
And these robust inputs have also been shown to produce features that \jmrv{are better aligned} with human perception and are easier to understand~\cite{santurkar2019image, tsipras2018robustness, ilyas2019adversarial}.

\begin{figure}[t]
\centering
\includegraphics[width=0.85\linewidth]{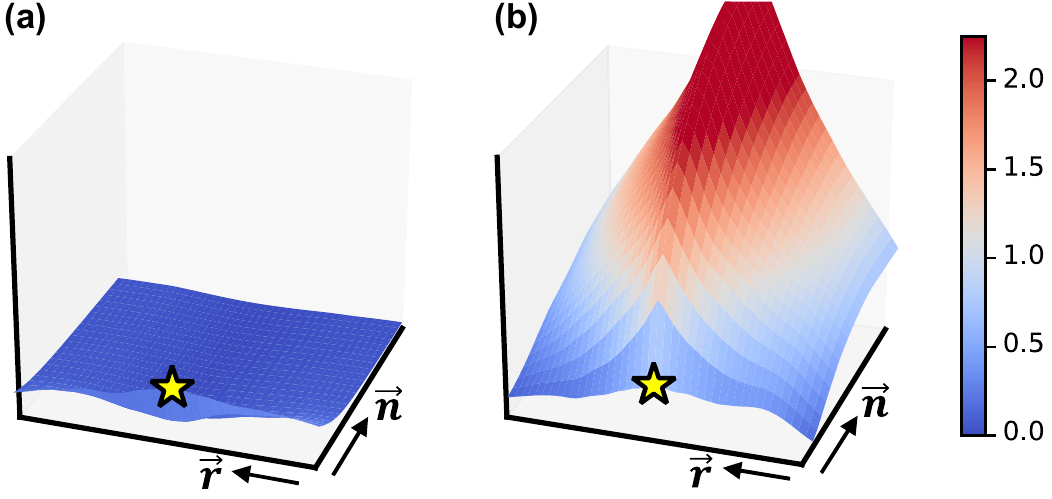}
\vspace{-1.4em}
\caption{\label{landscape} Loss landscapes obtained by manipulating images with weighted sums of the normal vector $\vec{n}~$ and a random vector $\vec{r}~$ for (a) adversarial climbing and (b) adversarial attack. The yellow star corresponds to $x^0$.}
\vspace{-.6em}
\end{figure}

We can therefore expect that images manipulated by adversarial climbing will similarly produce features that \jmrv{are aligned} with human perception, because adversarial climbing drives the input towards a flatter \rv{region} of the loss landscape.
We support this assertion \rv{by visualizing} the loss landscape of our trained classifier (Fig.~\ref{landscape}), following Moosavi \textit{et al.}~\cite{moosavi2019robustness}. We obtain a normal (manipulation) vector $\vec{n}$ from the classification loss $\ell$ computed from an image, and a random vector $\vec{r~}$. 
We then plot classification loss values computed from manipulated images using vectors \rvt{that} are interpolated between $\vec{n}$ and $\vec{r}$ using a range of interpolation ratios.
Inputs perturbed by adversarial climbing (Fig.~\ref{landscape}(a)) lie on a flatter loss landscape than those perturbed by an adversarial attack (Fig.~\ref{landscape}(b)).
Therefore, it is fair to expect that adversarial climbing will enhance the \rvt{class-relevant features} from the human viewpoint.

\begin{figure}[t]
\centering
\includegraphics[width=\linewidth]{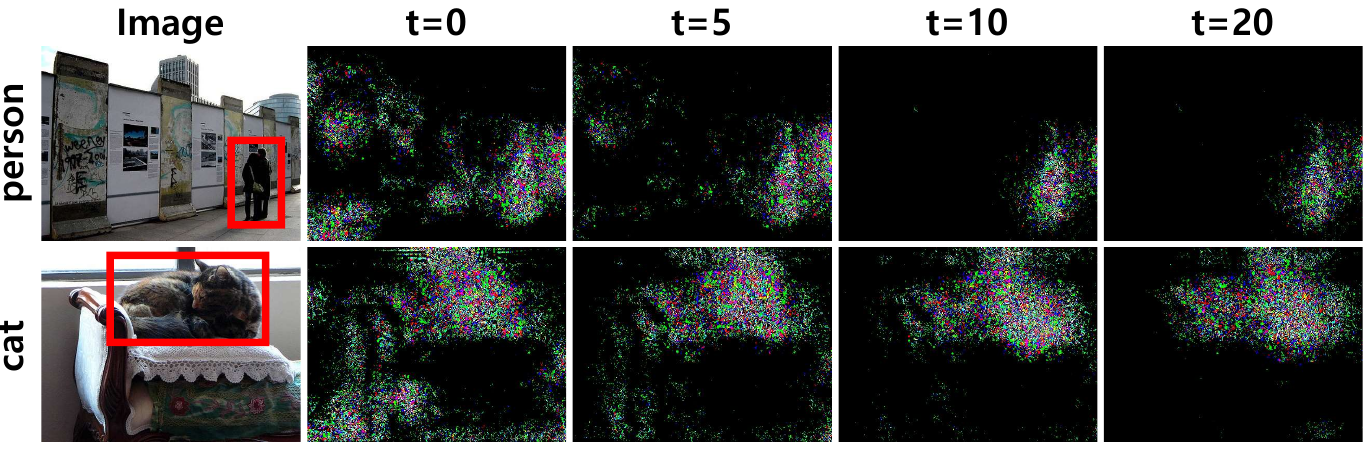}
\vspace{-2.3em}
\caption{\label{grad_sample} \rv{Examples of visual saliency maps~\cite{simonyan2013deep}} of manipulated images at iterations $t=0$, $t=5$, $t=10$, and $t=20$. The red box represents the target object.
}
\vspace{-1em}
\end{figure}

We provide further empirical evidence in support of this assertion. 
Fig.~\ref{grad_sample} shows saliency maps produced by a classifier from values of $|\nabla_{x^{t}} \texttt{NN}(x^{t}) / \max \nabla_{x^{t}} \texttt{NN}(x^{t})|$ at each iteration $t$.
These maps show the regions of an image
that were particularly influential in the classification~\cite{zeiler2014visualizing, simonyan2013deep}. 
When $t=0$, the gradients are very noisy, but as $t$ increases, the map identifies the target object more clearly. 
We infer that, as adversarial climbing progresses, the classifier increasingly focuses on the regions which are regarded as class-relevant from the human viewpoint.

\subsection{Regularization}\label{reg_sec}
\rv{If} adversarial climbing is \rv{performed to excess}, regions corresponding to objects \rv{in the wrong classes} may be activated, or the attribution scores of regions that already have high scores may be unintentionally increased.
We address these issues by (\RNum{1}) suppressing the logits associated with classes other than the target class, and (\RNum{2}) restricting the attribution scores of discriminative regions, which already have high scores. 

\begin{figure}[t]
\centering
\includegraphics[width=0.88\linewidth]{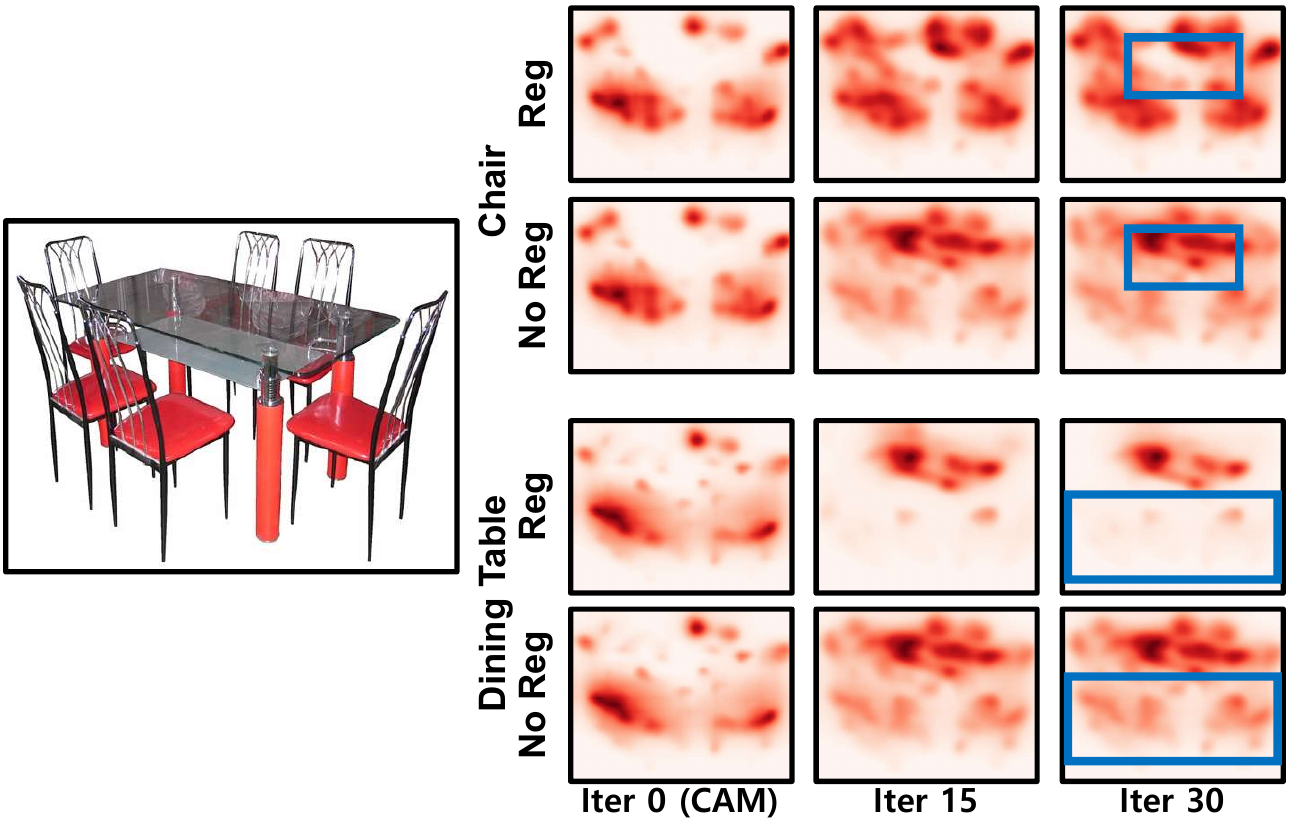}
\vspace{-1em}
\caption{\label{ex_supp_other_classes} Examples of localization maps (\rv{$\mathcal{A}$}) of the `chair' and `dining table', with and without regularization for suppressing other classes.}
% Effectiveness of regularization technique. (\textit{Left}) Seed quality on mIoU (\%). (\textit{Right}) 1-precision (\%)}
\vspace{-1.5em}
\end{figure}

\textbf{Suppressing Other Classes:}
Several objects of different classes in an image can simultaneously influence one another's \rv{logits}.
Thus, even if adversarial climbing increases the logit of the target class, it can also \rv{\rvt{increase} the attribution scores of} regions \rvt{corresponding to} objects of other classes that frequently appear together \rv{with objects of the target class} in an image.
For example, a chair and a dining table often \rv{appear} together in an image, and \rv{therefore}, increasing the logit \rv{of the} chair by adversarial climbing can cause regions corresponding to the dining table to be incorrectly identified (the blue boxes in Fig.~\ref{ex_supp_other_classes}). 
Therefore, we perform regularization to reduce \rv{the logits of all the classes} except the target class $c$\rv{; this regularization process} allows only the region corresponding to the target class to be activated, as shown in Fig.~\ref{ex_supp_other_classes}.

\begin{figure}[t]
\centering
\includegraphics[width=\linewidth]{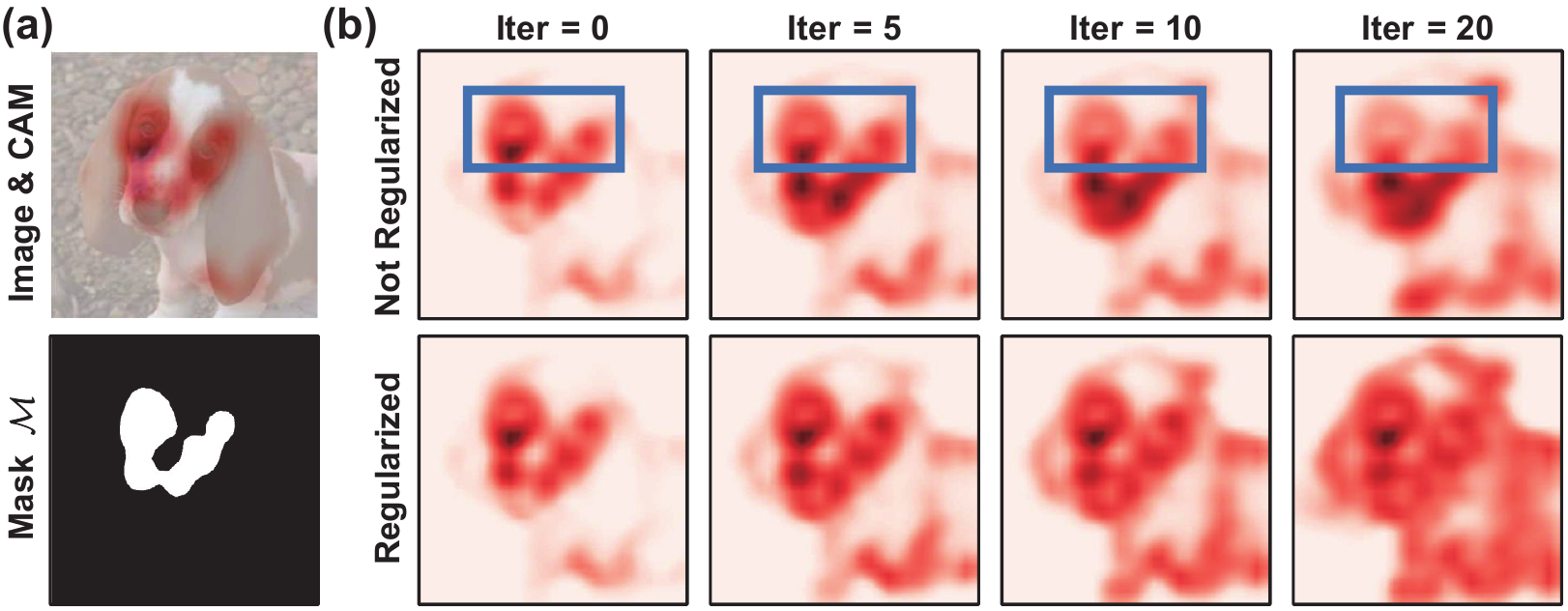}
\vspace{-1.9em}
\caption{\label{mask_ex} (a) An example image with its CAM and restricting mask $\mathcal{M}$. (b) The initial CAM, and localization maps (\rv{$\mathcal{A}$}) after 5, 10, and 20 steps of adversarial climbing, with and without regularization.}
% \vspace{-1.em}
\end{figure}

\textbf{Restricting High Attributions:}
As shown in Fig.~\ref{fig_amp}(a), adversarial climbing increases the attribution scores of both discriminative and non-discriminative regions \rv{in} the feature map.
The excessive growth of attribution scores \rv{associated with} discriminative regions is problematic for two reasons: it prevents new regions from being attributed to the classification score; and, if \rv{the maximum attribution score} increases during adversarial climbing, the normalized scores of the remaining \rv{regions} \rv{are likely to} decrease (blue boxes in Fig.~\ref{mask_ex}(b)).

Therefore, during adversarial climbing, we limit the growth \rvt{of} attribution scores in regions \rv{already with high scores}, so that the attribution scores of those regions remain similar to those of $x^{0}$.
We realize this scheme by introducing a restricting mask $\mathcal{M}$ that contains the regions with attribution scores in $\texttt{CAM}(x^{t-1})$ that are higher than \rv{a} threshold $\tau$. 
The restricting mask $\mathcal{M}$ can be represented as follows:
\begin{equation}\label{mask}
\mathcal{M} = \mathbbm{1}(\texttt{CAM}(x^{t-1}) > \tau),
\end{equation}
where $\mathbbm{1}(\mathord{\cdot})$ is an indicator function. An example mask $\mathcal{M}$ is shown in Fig.~\ref{mask_ex}(a). 

We add a regularization term so that the values in $\texttt{CAM}(x^{t-1})$ \rv{which} correspond to the regions of $\mathcal{M}$ are forced to be equal to \rv{those of} same regions in $\texttt{CAM}(x^{0})$. 
This regularization keeps $s^{i \in {R_\text{D}}}_t$ fairly constant, but $s^{i \in {R_\text{ND}}}_t$ still grows during adversarial climbing (Fig.~\ref{fig_amp}(b)).
Fig.~\ref{fig_amp} shows that adversarial climbing enhances non-discriminative features \rvt{rather} than discriminative features (by a factor of less than 2), and regularization magnifies this difference (to a factor of 2.5 or more).
As a result, new regions of the target object are found more effectively, as shown in Fig.~\ref{mask_ex}(b).

The two regularization terms introduced above modify Eq.~\ref{sgd} as follows:

\vspace{-0.5em}
\begin{equation}\label{sgd_reg}
x^{t} = x^{t-1} + \xi \nabla_{x^{t-1}} \mathcal{L}, ~~\text{where}
\end{equation}
\begin{align}\label{reg_loss}
\begin{split}
\mathcal{L} = &~ y_c^{t-1} - \sum_{k \in \mathcal{C} \setminus c}  y_k^{t-1} -  \lambda \left\lVert \mathcal{M} \odot |\texttt{CAM}(x^{t-1}) -  \texttt{CAM}(x^{0})|\right\lVert_1.
% \vspace{-0.5em}
\end{split}
\end{align}
$\mathcal{C}$ is the set of all classes, $\lambda$ is a hyper-parameter that controls the influence of the regularizing mask, and $\odot$ is element-wise multiplication.

\subsection{External saliency supervision}\label{method_sal}
In weakly supervised semantic segmentation, we can optionally utilize a salient object detector~\cite{hou2017deeply, jiang2013salient, liu2019simple}, which provides boundary information about the salient objects in an image. Since it is difficult to obtain the exact boundary of the target object using \rv{\rvt{the} image label alone}, a salient object detector is very useful \rv{in} weakly supervised semantic segmentation.
A salient object detector \rvt{that} works in a class-agnostic manner meets the basic assumptions of weakly supervised semantic segmentation. Many previous \rv{methods of} weakly supervised semantic segmentation~\cite{sun2020mining, lee2019ficklenet, lee2019frame, zhangsplitting, fan2018cian, fanemploying, wei2017object, wei2018revisiting} have used a salient object detector, but \rvt{its use has generally been limited} to the post-processing step, \rv{in which} explicit background cues \rv{are obtained}, and the initial seed \rv{is refined}.

We seamlessly integrate a salient object detector \rv{into} adversarial climbing without significant modifications to our method.
We use it to prevent excessive expansion of the CAM during adversarial climbing.
With the integration of the salient object detector, Eq.~\ref{mask} is modified as:
$\mathcal{M} = \mathbbm{1}(\texttt{CAM}(x^{t-1}) > \tau) \cup \mathcal{D}$, where $\mathcal{D}$ denotes the \rv{region} of an image identified as background by the salient object detector.
This new $\mathcal{M}$ is then used in Eq.~\ref{reg_loss}. It restricts any \rvt{further} increase in the attribution scores of the region $\mathcal{D}$ during adversarial climbing, so that the \rv{attribution scores of regions} outside the object \rvt{do not increase}.

\subsection{Weakly Supervised Semantic Segmentation}\label{train_segnet}
\textbf{Generating pseudo ground truth:} Since a CAM is obtained from down-sampled intermediate features produced by the classifier, it localizes the target object coarsely and cannot represent its exact boundary.
\rv{Therefore, 
many weakly supervised semantic segmentation methods~\cite{chang2020weakly, chang2020mixup, wang2020self, zhang2020causal, liu2020weakly} regard their localization map as \rv{an} initial seed, and construct pseudo ground truths by modifying their initial seeds using established \rv{refinement} methods~\cite{huang2018weakly, ahn2018learning, ahn2019weakly}.} 
Similarly, we obtain \rv{pseudo ground truths} by applying seed refinement to the coarse map $\mathcal{A}$. 
We can further refine the resulting pseudo ground \rv{truths} with a saliency map obtained from a salient object detector.
When a foreground pixel in a pseudo label is marked as background on \rv{this saliency} map, or a background pixel is marked as foreground, we change the pseudo label of that pixel to `ambiguous'.
Regions labeled as `ambiguous' are subsequently ignored in the training of the segmentation network.

\textbf{Training segmentation networks:} In weakly supervised learning, we use the pseudo ground truth\rv{s obtained in the manner described above} for training DeepLab-v2, pre-trained on the \rvt{ImageNet-1K} dataset~\cite{deng2009imagenet}. 
For semi-supervised learning, we employ CCT~\cite{ouali2020semi}, which uses IRN~\cite{ahn2019weakly} to generate pseudo ground truth masks; we replace these with our masks, constructed as described above.

\subsection{Weakly Supervised Object Localization}\label{wsol_method}
Adversarial climbing can also be extended to weakly supervised
object localization.
\rv{This extension demonstrates the generality of our method in two aspects:
1) The datasets used for weakly supervised semantic segmentation and object localization differ significantly.
The datasets used for weakly supervised object localization contain images intended for use in fine-grained classification (CUB-200-2011~\cite{wah2011caltech}) or images with 1,000 object classes (ImageNet-1K~\cite{deng2009imagenet}), which is far more diverse than that used for weakly supervised semantic segmentation, which usually involves between 20 or 80 classes.
2) It allows us to compare \rvt{the localization capability of our method against that of other recent weakly supervised object localization methods,} which share the same goal with us.
}

We apply \rv{adversarial} climbing to the baseline methods CAM~\cite{zhou2016learning} and CutMix~\cite{yun2019cutmix} because they use a vanilla classifier. 
We manipulate an input image by adversarial climbing and obtain $\mathcal{A}$, as described in Sections~\ref{Advcam_method} and \ref{reg_sec}. We do not use seed refinement or a salient object detector \rv{in this context} to \rv{allow for} a fair comparison with other methods.
We then generate bounding boxes from $\mathcal{A}$, following Choe \textit{et al.}~\cite{choe2020evaluating}.

For all \rv{the} hyper-parameters except $\lambda$, we use the same settings \rv{as those used for} adversarial climbing with weakly supervised semantic segmentation. 
The classifiers used in weakly supervised semantic segmentation and weakly supervised object localization are trained using different loss functions: sigmoid-based cross-entropy loss (multi-label classification) was used for the former, and softmax-based cross-entropy loss (single-label classification) for the latter. 
Since these two loss functions yield different distributions of classification logits, the value of $\lambda$ is adjusted for each task (but not between datasets).

\section{Experiments on Semantic Segmentation}
\subsection{Experimental Setup}\label{setup_sec}

\textbf{Datasets:} We conducted experiments on the PASCAL VOC 2012~\cite{everingham2010pascal} and MS COCO 2014~\cite{lin2014microsoft} datasets.
The images in these datasets come with masks for fully supervised semantic segmentation, but we only used these masks for evaluation. 
The PASCAL VOC dataset, as augmented by Hariharan \textit{et al.}~\cite{hariharan2011semantic}, contains 10,582 training images, depicting objects of 20 classes. In a weakly supervised setting, we trained our network on all 10,582 training images using their class labels. In a semi-supervised setting, we used 1,464 training images \rvt{that} have pixel-level annotations and the remaining 9,118 training images with \rv{their} class labels, following previous work~\cite{lee2019ficklenet, ouali2020semi, wei2018revisiting, luosemi}.
The MS COCO 2014 dataset has 80K training images depicting objects of 80 classes.
We evaluated our method on 1,449 validation images and 1,456 test images from \rv{PASCAL VOC 2012} and on 40,504 validation images from \rv{MS COCO 2014}, by calculating mean intersection-over-union (mIoU) values.

\textbf{Reproducibility:}
For both the \rv{PASCAL VOC 2012} and \rv{MS COCO 2014} datasets, we performed iterative adversarial climbing with $T=27$ and $\xi=0.008$. We set $\lambda$ to 7 and $\tau$ to 0.5.
To generate the initial seed, we followed the procedure of Ahn \textit{et al.}~\cite{ahn2019weakly}, including the use of ResNet-50~\cite{he2016deep}. 
For final segmentation, we used DeepLab-v2-ResNet101~\cite{chen2017deeplab} as the backbone network. 
We used the default settings of DeepLab-v2~\cite{chen2017deeplab} in training \rv{with} the \rv{PASCAL VOC 2012} dataset.
For the \rv{MS COCO 2014} dataset, we cropped the training images to 481$\times$481 pixels, rather than 321$\times$321, \rv{to make better use of the larger images in this dataset}.
We used \rv{the} salient object detector provided by Hou \textit{et al.}~\cite{hou2017deeply}, following previous work~\cite{li2020group, yao2021nonsalient}.
In a semi-supervised setting, we used the same \rv{setup} as Ouali \textit{et al.}~\cite{ouali2020semi}, including the ResNet-50 backbone. \rv{This does not include a \rvt{salient} object detector.}
\begin{figure*}[t]
\centering
\includegraphics[width=\linewidth]{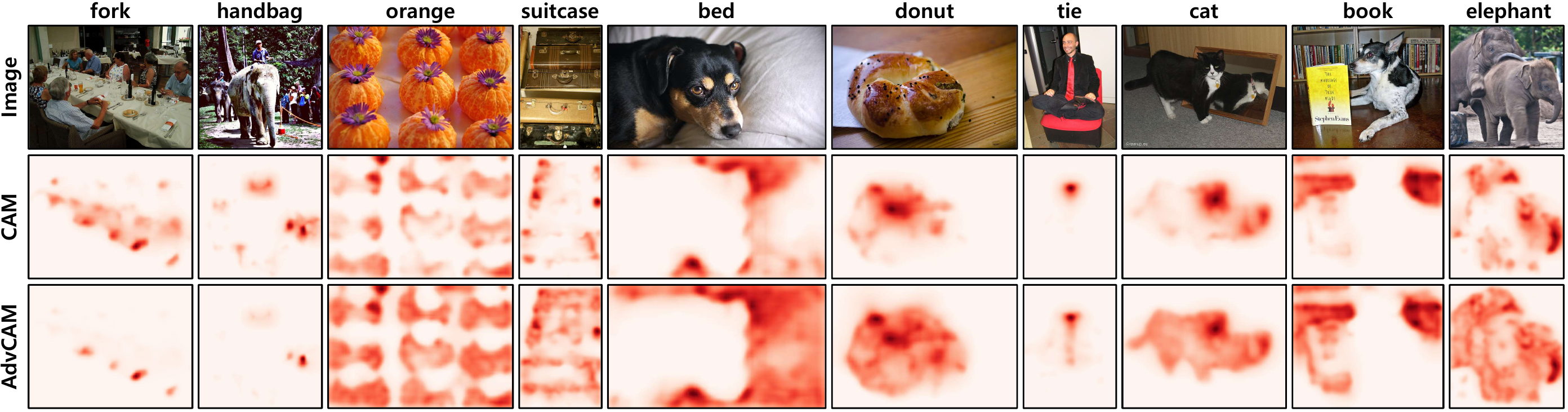}
\vspace{-1.7em}
\caption{\label{cococamsamples} Examples of \rv{the final localization maps ($\mathcal{A}$)} obtained from the CAM and AdvCAM of the \rv{MS COCO 2014} dataset.}
\vspace{-.6em}
\end{figure*}

\begin{table}[tbp]
\normalsize
  \centering
  \caption{mIoU (\%) of the initial seed (Seed), the seed with CRF (+CRF), and the pseudo ground truth mask (Mask) on PASCAL VOC 2012 \textit{train} images.}
  \vspace{-0.7em}
    \begin{tabular}{l@{\hskip 0.3in}c@{\hskip 0.1in}c@{\hskip 0.08in}c}
     \Xhline{1pt}\\[-0.95em]
    Method  &  Seed  & + CRF & Mask \\
    \hline\hline \\[-0.9em]
    \multicolumn{4}{l}{Seed Refine with PSA~\cite{ahn2018learning}:} \\
    $\text{PSA}_{\text{~~CVPR '18}}$~\cite{ahn2018learning} & 48.0 & - & 61.0 \\
    $\text{Mixup-CAM}_{\text{~~BMVC '20}}$~\cite{chang2020mixup} & 50.1 & - & 61.9 \\
    $\text{Chang \textit{et al.}}_{\text{~~CVPR '20}}$~\cite{chang2020weakly} & 50.9 & 55.3 & 63.4 \\
    $\text{SEAM}_{\text{~~CVPR '20}}$~\cite{wang2020self}  & 55.4 & 56.8& 63.6 \\
    AdvCAM (Ours) & \textbf{55.6}  & \textbf{62.1} & \textbf{68.0}\\
    % $\text{CAM + BEM}_{\text{~~ECCV '20}}$~\cite{chenweakly}  & 50.4 & 67.2\\
    \hline\\[-0.9em]
    \multicolumn{4}{l}{Seed Refine with IRN~\cite{ahn2019weakly}:} \\
    $\text{IRN}_{\text{~~CVPR '19}}$~\cite{ahn2019weakly} & 48.8  & 54.3 & 66.3 \\
    $\text{MBMNet}_{\text{~~ACMMM '20}}$~\cite{liu2020weakly} & 50.2 & - & 66.8 \\
    $\text{CONTA}_{\text{~~NeurIPS '20}}$~\cite{zhang2020causal} & 48.8 & - & 67.9 \\

    AdvCAM (Ours) & \textbf{55.6}  & \textbf{62.1} & \textbf{69.9} \\
    \hline\\[-0.9em]
    \multicolumn{4}{l}{\rv{Utilization of saliency supervision:}} \\
    $\text{EPS}_{\text{~~CVPR '21}}$~\cite{lee2021railroad} & - & - & \rv{71.6} \\
    AdvCAM--Sal (Ours) & \textbf{60.8}  & \textbf{66.6} & \textbf{72.3} \\

    \Xhline{1pt}
    \vspace{-1em}
    \end{tabular}%
  \label{table_seed}%
\end{table}%

\begin{table}[t]
\normalsize
  \centering
  \caption{Improvement in IRN~\cite{ahn2019weakly} achieved by CONTA~\cite{zhang2020causal} and our method, in terms of the mIoU (\%) of the initial seed (Seed), and the pseudo ground truth mask (Mask) on \rv{MS COCO 2014} \textit{train} images. }
  \vspace{-1em}
%   Table A1: Performance improvement of IRN [\textcolor{green}{2}] by ours and CONTA~[\textcolor{green}{61}] for the seed and the pseudo mask (Mask) on COCO \textit{train} set, and the segmentation (Seg) on COCO \textit{val} set.}
    \begin{tabular}{l@{\hskip 0.2in}c@{\hskip 0.15in}c}
     \Xhline{1pt}\\[-1.em]
    Method  &  Seed  & Mask \\ \\[-1.15em]
    \hline\hline\\[-0.9em]
    IRN~\cite{ahn2019weakly} $\rightarrow$ CONTA~\cite{zhang2020causal} & 27.4 $\rightarrow$ 28.7 & 34.0 $\rightarrow$ 35.2  \\

    IRN~\cite{ahn2019weakly} $\rightarrow$ Ours & 33.5 $\rightarrow$ 37.2  & 42.9 $\rightarrow$ 46.0  \\
    \\[-1.2em]
    \Xhline{1pt}
    \vspace{-2em}
    \end{tabular}%
    
  \label{coco_seed_table}%
\end{table}%

\subsection{Quality of the Initial Seed}\label{initialseedexp}
\textbf{Results on \rv{PASCAL VOC 2012}.} Table~\ref{table_seed} compares, in terms of mIoU, the initial seed and pseudo ground truth masks obtained by our method and by other recent techniques. Both seeds and masks were generated from training images from the \rv{PASCAL VOC 2012} dataset. 
We determined the best initial seeds by applying a range of thresholds to separate the foreground and background in the map $\mathcal{A}$, following SEAM~\cite{wang2020self}.
Our initial seeds are 6.8\%p better than \rv{the baseline provided by} the original CAMs~\cite{zhou2016learning}, and our seeds outperform those produced by other methods.
In particular, we obtained better initial seeds than SEAM~\cite{wang2020self}, which uses an auxiliary self-attention module to perform pixel-level refinement of the initial CAM by considering the relationship between pixels. 

\begin{table}[t]
\centering
\normalsize
  \caption{Weakly supervised semantic segmentation performance on PASCAL VOC 2012 \textit{val} and \textit{test} images. \rv{All the methods use ResNet-based backbones~\cite{he2016deep}.}}\label{table_semantic}
%   Level of supervision: $\mathcal{F}-$full, $\mathcal{I}-$image label, $\mathcal{B}-$box, $\mathcal{S}-$saliency.}  
\vspace{-0.7em}
\begin{threeparttable}
\begin{tabular}{l@{\hskip 0.3in}c@{\hskip 0.3in}cc}
    \Xhline{1pt}\\[-0.95em]
    Method  & Sup.& \textit{val} & \textit{test}\\
    \hline\hline 
    \\[-0.9em]
    
    \multicolumn{4}{l}{Supervision: Stronger than image labels} \\
    $\text{DeepLab}_{\text{~~TPAMI '17}}$~\cite{chen2017deeplab}  & $\mathcal{F}$  & 76.8  & 76.2 \\
    $\text{SDI}_{\text{~~CVPR '17}}$~\cite{khoreva2017simple}   & $\mathcal{B}$ & 69.4  & -  \\
    $\text{Song \textit{et al.}}_{\text{~~CVPR '19}}$~\cite{song2019box}   & $\mathcal{B}$ & 70.2  & - \\
    $\text{BBAM}_{\text{~~CVPR '21}}$~\cite{lee2021bbam}   & $\mathcal{B}$ & 73.7  & 73.7 \\
    % $\text{Box2Seg}_{\text{~~ECCV '20}}$~\cite{kulharia12356box2seg}    & 76.4  & - \\
        \\[-0.9em]
\hline
    \\[-0.9em]
    \multicolumn{3}{l}{Supervision: Image-level tags}\\
    $\text{IRN}_{\text{~~CVPR '19}}$~\cite{ahn2019weakly}  &  $\mathcal{I}$ & 63.5 & 64.8 \\
    $\text{SSDD}_{\text{~~ICCV '19}}$~\cite{Shimoda_2019_ICCV}    & $\mathcal{I}$   & 64.9  & 65.5\\
    $\text{SEAM}_{\text{~~CVPR '20}}$~\cite{wang2020self}    & $\mathcal{I}$ & 64.5  & 65.7 \\
    % $\text{Mixup-CAM}_{\text{~~BMVC '20}}$~\cite{chang2020mixup}   & $\mathcal{I}$  & 65.6  & -\\
    $\text{Chen \textit{et al.}}_{\text{~~ECCV '20}}$~\cite{chenweakly}   &   $\mathcal{I}$ & 65.7  & 66.6  \\

    $\text{Chang \textit{et al.}}_{\text{~~CVPR '20}}$~\cite{chang2020weakly}   & $\mathcal{I}$  & 66.1  & 65.9\\
    % $\text{RRM}_{\text{~~AAAI '20}}$~\cite{zhang2019reliability}     & 66.3  & 66.5   \\
    
    $\text{CONTA}_{\text{~~NeurIPS '20}}$~\cite{zhang2020causal}   & $\mathcal{I}$  & 66.1  & 66.7  \\

    AdvCAM (Ours) & $\mathcal{I}$ & \textbf{68.1} & \textbf{68.0}  \\
    \\[-0.9em]
\hline
    \\[-0.9em]
    \multicolumn{3}{l}{Supervision: Image-level tags + Saliency}\\
%     MIL-FCN (ICLR '15) &   10K    & 25.7  & 24.9 \\
%     CCNN (ICCV '15) & 10K   & 35.3  & 35.6 \\
%     EM\_Adapt (ICCV '15) & 10K   & 38.2  & 39.6 \\
%     $\text{DCSM}_{\text{~~ECCV '16}}$~\cite{shimoda2016distinct} & 10K   & 44.1  & 45.1 \\
%     $\text{BFBP}_{\text{~~ECCV '16}}$~\cite{saleh2016built} & 10K   & 46.6  & 48.0 \\
%     $\text{SEC}_{\text{~~ECCV '16}}$~\cite{kolesnikov2016seed}    & 50.7  & 51.1 \\
% %     $\text{Saleh et al.}_{\text{~~TPAMI '17}}$~\cite{saleh2018incorporating} & 10K   & 50.9  & 52.6 \\
%     $\text{CBTS-cues}_{\text{~~CVPR '17}}$~\cite{roy2017combining}    & 52.8  & 53.7 \\
%     $\text{TPL}_{\text{~~ICCV '17}}$~\cite{kim2017two}   & 53.1  & 53.8 \\
%     $\text{AE\_PSL}_{\text{~~CVPR '17}}$~\cite{wei2017object}    & 55.0    & 55.7 \\
%     $\text{DCSP}_{\text{~~BMVC '17}}$~\cite{chaudhry2017discovering}  & 58.6 & 59.2\\
%     $\text{MEFF}_{\text{~~CVPR '18}}$~\cite{ge2018multi}  & - & 55.6\\
    % $\text{GAIN}_{\text{~~TPAMI '19}}$~\cite{li2019guided}    & 59.4  & 59.6 \\
    % $\text{MCOF}_{\text{~~CVPR '18}}$~\cite{wang2018weakly}    & 60.3  & 61.2 \\
    % $\text{DSRG}_{\text{~~CVPR '18}}$~\cite{huang2018weakly}   & 61.4   & 63.2 \\
    % $\text{AffinityNet}_{\text{~~CVPR '18}}$~\cite{ahn2018learning}    & 61.7  & 63.7 \\
    
    $\text{SeeNet}_{\text{~~NeurIPS '18}}$~\cite{hou2018self}  & $\mathcal{I}$, $\mathcal{S}$ & 63.1 & 62.8 \\
    $\text{Li \textit{et al.}}_{\text{~~ICCV '19}}$~\cite{li2019attention}   &   $\mathcal{I}$, $\mathcal{S}$ & 62.1  & 63.0  \\
    $\text{FickleNet}_{\text{~~CVPR '19}}$~\cite{lee2019ficklenet}  & $\mathcal{I}$, $\mathcal{S}$ & 64.9 & 65.3\\
    $\text{Lee \textit{et al.}}_{\text{~~ICCV '19}}$~\cite{lee2019frame}   & $\mathcal{I}$, $\mathcal{S}$, $\mathcal{W}$  & 66.5  & 67.4  \\
    % $\text{OAA+}_{\text{~~ICCV '19}}$~\cite{JiangOAAICCV19}     & 65.2  & 66.4 \\
    $\text{CIAN}_{\text{~~AAAI '20}}$~\cite{fan2018cian}    & $\mathcal{I}$, $\mathcal{S}$ & 64.3  & 65.3 \\
    $\text{Zhang \textit{et al.}}_{\text{~~ECCV '20}}$~\cite{zhangsplitting}   & $\mathcal{I}$, $\mathcal{S}$  & 66.6  & 66.7  \\
    $\text{Fan \textit{et al.}}_{\text{~~ECCV '20}}$~\cite{fanemploying}   & $\mathcal{I}$, $\mathcal{S}$  & 67.2  & 66.7  \\
    $\text{Sun \textit{et al.}}_{\text{~~ECCV '20}}$~\cite{sun2020mining}   & $\mathcal{I}$, $\mathcal{S}$  & 66.2  & 66.9  \\
    
    $\text{LIID}_{\text{~~TPAMI '20}}$~\cite{liu2020leveraging}    & $\mathcal{I}$, $\mathcal{S}$ & 66.5  & 67.5 \\
    $\text{Sun \textit{et al.}}_{\text{~~ECCV '20}}$~\cite{sun2020mining}   & $\mathcal{I}$, $\mathcal{S}$, $\mathcal{W}$  & 67.7  & 67.5  \\
    $\text{Li \textit{et al.}}_{\text{~~AAAI '21}}$~\cite{li2020group}   & $\mathcal{I}$, $\mathcal{S}$  & 68.2  & 68.5  \\
    $\text{Yao \textit{et al.}}_{\text{~~CVPR '21}}$~\cite{yao2021nonsalient}   & $\mathcal{I}$, $\mathcal{S}$  & 68.3  & 68.5  \\
    % \rv{$\text{LIID}_{\text{~~TPAMI '20}}$~\cite{liu2020leveraging}$^\dagger$} & \rv{$\mathcal{I}$, $\mathcal{S}$} & \rv{{69.4}} & \rv{{70.4}}  \\
    % \rv{$\text{A$^2$GNN}_{\text{~~TPAMI '21}}$~\cite{zhang2021affinity}$^\dagger$} & \rv{$\mathcal{I}$, $\mathcal{S}$} & \rv{{69.0}} & \rv{{69.6}}  \\
    % $\text{DRS}_{\text{~~AAAI '21}}$~\cite{kim2021discriminative}   & $\mathcal{I}$, $\mathcal{S}$  & 71.2  & 71.4  \\
    % $\text{SPML}_{\text{~~ICLR '21}}$~\cite{ke2021universal}   & $\mathcal{I}$, $\mathcal{E}$  & 70.6  & 71.5  \\
    \rv{$\text{A$^2$GNN}_{\text{~~TPAMI '21}}$~\cite{zhang2021affinity}}   & \rv{$\mathcal{I}$, $\mathcal{S}$}  &   \rv{68.3} & \rv{68.7}\\
    $\text{RIB}_{\text{~~NeurIPS '21}}$~\cite{lee2021reducing}    & $\mathcal{I}$, $\mathcal{S}$ & 70.2  & 70.0 \\
    AdvCAM--Sal (Ours) & $\mathcal{I}$, $\mathcal{S}$ & \textbf{71.3} & \textbf{71.2} \\
    % \rv{AdvCAM--Sal (Ours)$^\dagger$} & \rv{$\mathcal{I}$, $\mathcal{S}$} & \rv{\textbf{72.0}} & \rv{\textbf{72.7}}  \\
    % \rv{AdvCAM--Sal (Ours)$^\ddagger$} & \rv{$\mathcal{I}$, $\mathcal{S}$} & \rv{\textbf{73.0}} & \rv{\textbf{72.7}}  \\

    % \\[-0.9em]
    \Xhline{1pt}
    
    \end{tabular}%
    \begin{tablenotes}
  \footnotesize
\item $\mathcal{F}-$full, $\mathcal{I}-$image class, $\mathcal{B}-$box, $\mathcal{S}-$saliency, $\mathcal{W}-$web\\
\scriptsize
% $^*$ Anonymous result link can be found \href{http://host.robots.ox.ac.uk:8080/anonymous/TQSZTQ.html}{here}.
% $^*$\url{http://host.robots.ox.ac.uk:8080/anonymous/NE2KAJ.html}
        \end{tablenotes}
     \end{threeparttable}
    % \end{adjustbox}
    \vspace{-1em}

      \end{table}
\begin{table}[t]
\centering
\normalsize
\color{black}
  \caption{\rv{Weakly supervised semantic segmentation performance on PASCAL VOC 2012 \rvt{\textit{val}} and \rvt{\textit{test}} images using Res2Net-based backbones~\cite{gao2019res2net}. All the methods use saliency supervision.}}
%   Level of supervision: $\mathcal{F}-$full, $\mathcal{I}-$image label, $\mathcal{B}-$box, $\mathcal{S}-$saliency.}  
\begin{tabular}{lccc}
    \Xhline{1pt}\\[-0.95em]
    Method   & Backbone & \rvt{\textit{val}} & \rvt{\textit{test}} \\
    \hline\hline 
    \\[-0.9em]
    
    $\text{LIID}_{\text{~~TPAMI '20}}$~\cite{liu2020leveraging}  & Res2Net-101  &   69.4 & 70.4\\
    $\text{A$^2$GNN}_{\text{~~TPAMI '21}}$~\cite{zhang2021affinity}   & Res2Net-101  &   69.0 & 69.6\\

    AdvCAM--Sal (Ours)  &  Res2Net-101 &  72.0 & \textbf{72.7}\\
    AdvCAM--Sal (Ours)  &  Res2Net-152 &  \textbf{73.0} & \textbf{72.7} \\

    % \\[-0.9em]
    \Xhline{1pt}
    
    \end{tabular}%
      \label{res2net_result}%

      \end{table}

We applied a post-processing method for pixel refinement, based on a CRF, to the initial seeds produced by the method of Chang \textit{et al.}~\cite{chang2020weakly}, SEAM~\cite{wang2020self}, IRN~\cite{ahn2019weakly}, and our method.
CRF improved the seeds by more than 5\%p on average, except for the seeds from SEAM, which \rvt{is} only improved by 1.4\%p:
we believe \rv{that} this is because the seed from SEAM had already been refined by the self-attention module. 
After applying CRF, the seed produced by our method is 5.3\%p better than that from SEAM.

We also compared the pseudo ground truth masks extracted after seed refinement, with the masks produced by other methods, most of which refine their initial seeds with PSA~\cite{ahn2018learning} or IRN~\cite{ahn2019weakly}.
For a fair comparison, we produced pseudo ground truth masks using both of these seed refinement techniques. 
Table~\ref{table_seed} shows that our method outperforms the others by a large margin, independent of the seed refinement technique.

Finally, we assessed the quality of the initial seed and the pseudo ground truth mask obtained using the salient object detector introduced in Section~\ref{method_sal}\footnote{\rv{The results are obtained from the pseudo ground truths before introducing the ambiguous regions mentioned in Section~\ref{train_segnet}.}}. 
Table~\ref{table_seed} shows that this improved the initial seed and the pseudo ground truth mask by 5.2\%p and 2.4\%p respectively.
\rv{In addition, our method obtains better pseudo ground truth masks than EPS~\cite{lee2021railroad}, a recently introduced method \rvt{that} is considered to be state-of-the-art.}

\begin{figure*}[t]
\centering
\includegraphics[width=0.95\linewidth]{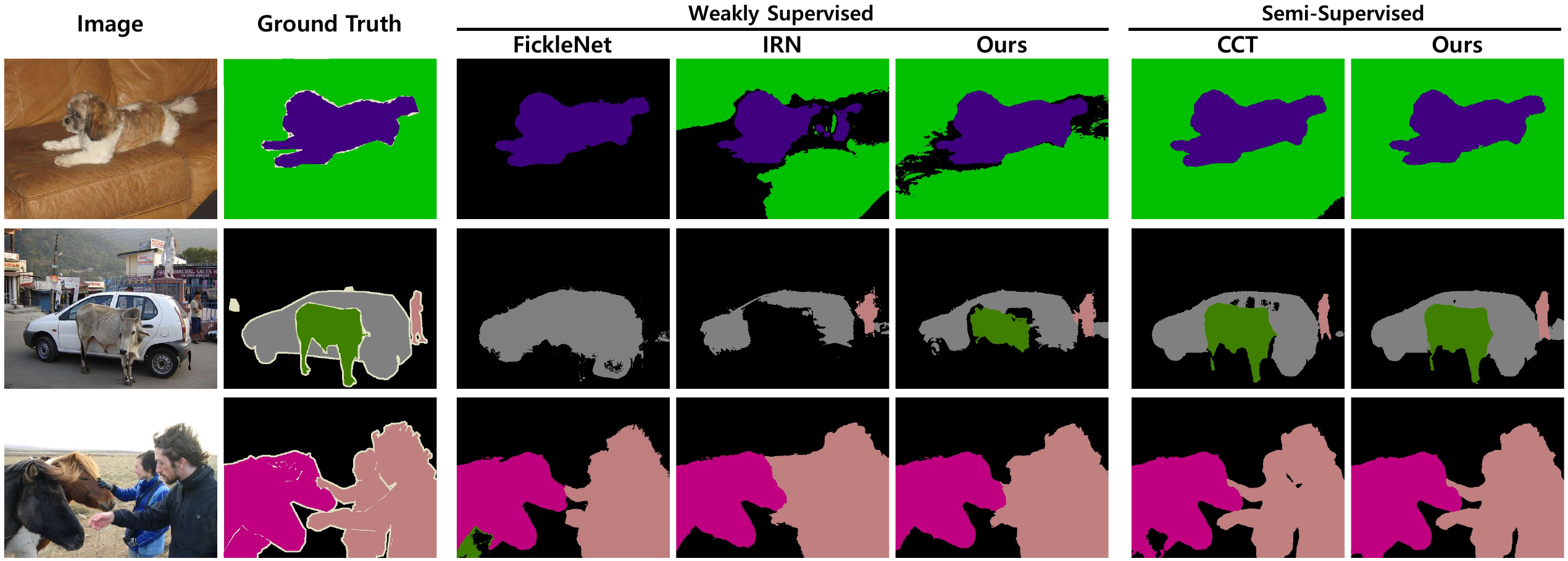}
\vspace{-.5em}
\caption{\label{segsample} Examples of predicted semantic masks for \rv{PASCAL VOC 2012} validation images from FickleNet~\cite{lee2019ficklenet}, IRN~\cite{ahn2019weakly}, and our method, with weak supervision; and for CCT~\cite{ouali2020semi} and our method in a semi-supervised context.}
% Effectiveness of regularization technique. (\textit{Left}) Seed quality on mIoU (\%). (\textit{Right}) 1-precision (\%)}
\vspace{-0.7em}
\end{figure*}

\begin{figure*}[t]
\centering
\includegraphics[width=0.95\linewidth]{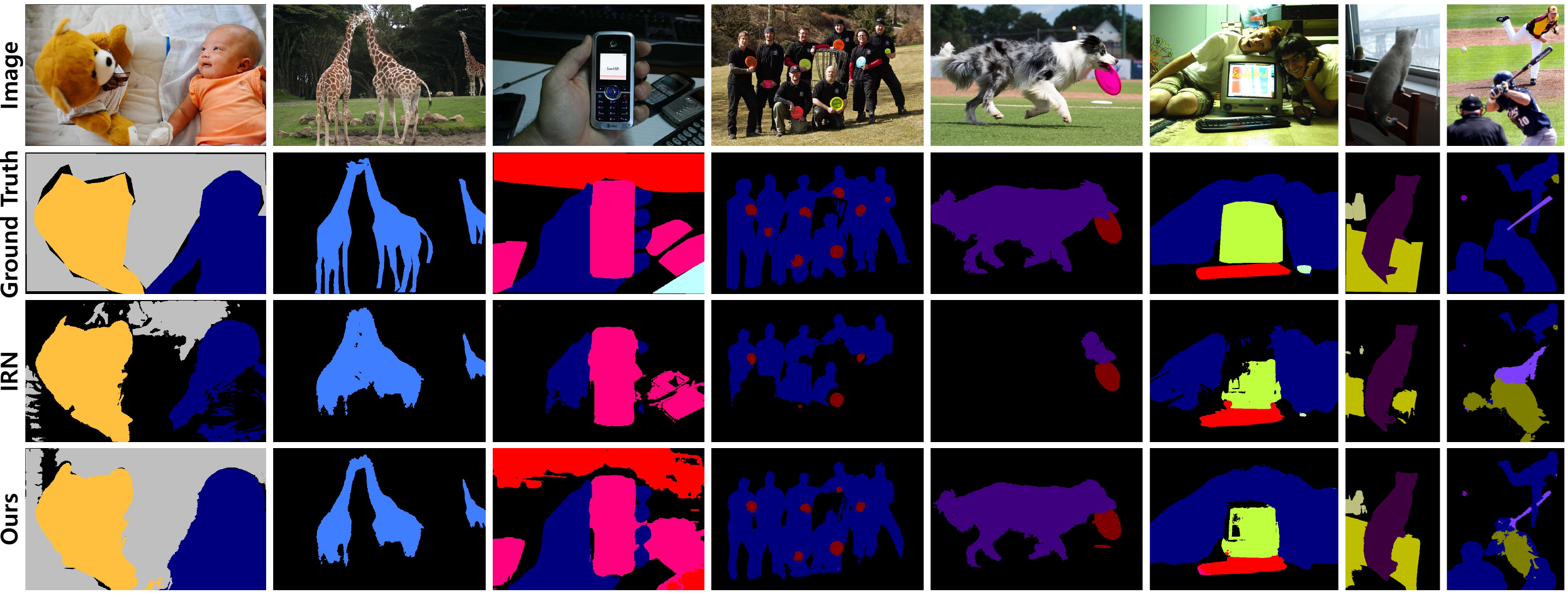}
\vspace{-.7em}
\caption{\label{segsample_coco} Examples of predicted semantic masks for \rv{MS COCO 2014} validation images from IRN~\cite{ahn2019weakly} and our method.}
% Effectiveness of regularization technique. (\textit{Left}) Seed quality on mIoU (\%). (\textit{Right}) 1-precision (\%)}
\vspace{-0.7em}
\end{figure*}

\textbf{Results on \rv{MS COCO 2014}.}
For \rv{MS COCO 2014} experiments, we implemented IRN~\cite{ahn2019weakly} using the official code. 
Since the results from IRN reported by Zhang \textit{et al.}~\cite{zhang2020causal} differ from \rv{those} that we obtained, we \rvt{compared} relative improvements.
Table~\ref{coco_seed_table} shows that our method improved the initial seed and pseudo ground truth masks by 3.7\%p and 3.1\%p respectively compared to IRN, while CONTA~\cite{zhang2020causal} \rv{improved them by} 1.3\%p and 1.2\%p respectively.
\rv{Fig.~\ref{cococamsamples} shows examples of the final localization maps obtained from the CAM and our method for MS COCO 2014 training images. These examples demonstrate that our method successfully \rv{identifies} more exact regions of the target object. 
For small objects (\textit{e.g.,} `fork' and `handbag' examples), the CAM~\cite{zhou2016learning} tends to mistakenly activate some \rv{regions} of the background, while our method mainly focuses on the \rv{region} corresponding to the target object.
For large objects, the CAM~\cite{zhou2016learning} is not able to \rv{identify all} of the target object, but our method can encompass them.
In short, our method obtains high-precision results for small objects and high-recall results for large objects. For a more detailed analysis and discussion, please refer to Section~\ref{classwise_analysis}.
}

\subsection{Weakly Supervised Semantic Segmentation}
\textbf{Results on \rv{PASCAL VOC 2012}.} Table~\ref{table_semantic} compares our method with other recently introduced weakly supervised semantic segmentation methods with various levels of supervision: fully supervised pixel-level masks \rv{($\mathcal{F}$)}, bounding boxes ($\mathcal{B}$) and image class labels ($\mathcal{I}$), with and without salient object masks ($\mathcal{S}$). All the results in Table~\ref{table_semantic} were obtained using a ResNet-based backbone~\cite{he2016deep}.
The results for the test images were obtained from the official \rv{PASCAL VOC 2012} evaluation server. 
With image-level annotation alone, our method achieved mIoU values of 68.1 and 68.0 for the PASCAL VOC 2012 validation and test images respectively. This is significantly better than the other methods under the same level of supervision. In particular, the mIoU value for validation images was 4.6\%p higher than that achieved by IRN~\cite{ahn2019weakly}, which is our baseline.
CONTA~\cite{zhang2020causal} performed best among our competitors, and achieved an mIoU value of 66.1; but CONTA uses SEAM~\cite{wang2020self}, which is known to outperform IRN~\cite{ahn2019weakly}. 
\rv{When CONTA is used in \rvt{conjunction} with IRN, it yields the mIoU value of 65.3, which is 2.8\%p worse than our method.}
Fig.~\ref{segsample} shows examples of semantic masks produced by FickleNet~\cite{lee2019ficklenet}, IRN~\cite{ahn2019weakly}, and our method.
\rv{These examples suggest that our method tends to capture the extent of the target object more exactly than previous methods. In addition, our method seldom misses a target object even in a complicated scene (\textit{e.g.,} the cow in the second row). More examples are presented in the Appendix.}

\rv{Even when our method was implemented only with class labels,} it outperformed other methods with auxiliary salient object mask supervision and those that require extra web images or videos~\cite{sun2020mining, lee2019frame}, except Li \textit{et al.}~\cite{li2020group} and Yao \textit{et al.}~\cite{yao2021nonsalient}, \rv{which are contemporaries of our method.} 
When we used saliency information, our segmentation results on the \rv{PASCAL VOC 2012} validation and test images were significantly better than those of Li \textit{et al.}~\cite{li2020group} and Yao \textit{et al.}~\cite{yao2021nonsalient}.
The performance of our method is also comparable to that of methods~\cite{song2019box, khoreva2017simple, lee2021bbam} based on bounding box supervision.
\rv{Table~\ref{res2net_result} compares our method with other recently introduced methods using Res2Net-based backbones~\cite{gao2019res2net}. Our method achieved \rv{an} mIoU value of 72.0 for the PASCAL VOC 2012 validation images, which is significantly better than that of the other methods under the same level of supervision \textit{i.e.,} image-class labels and saliency supervision. When \rv{the} more powerful Res2Net-152~\cite{gao2019res2net} is used, the performance of our method is further improved to 73.0 mIoU.}

\textbf{Results on \rv{MS COCO 2014}.}
Table~\ref{table_semantic_coco} compares weakly supervised semantic segmentation results from our method on the \rv{MS COCO 2014} dataset with \rvt{the results} from other recently introduced methods. Our method achieved an mIoU of 44.4 for the \rv{MS COCO 2014} validation images, which is significantly better than the mIoUs from the other methods. In particular, our method was 3.0\%p better than IRN~\cite{ahn2019weakly}, which is our baseline.

Fig.~\ref{segsample_coco} shows examples of predicted semantic masks for \rv{MS COCO 2014} validation images produced by IRN~\cite{ahn2019weakly} and by our method.
\rv{Because our initial seed covers the target object more precisely, it captures regions of the target object that IRN does not, leading to a more accurate boundary.
The last column of Fig.~\ref{segsample_coco} shows an example of the well-known class bias problem~\cite{li2019guided}, which is still largely open. The baseball player and the glove usually occur together, and as a result, the body of the player is sometimes mistakenly recognized as a glove. Nonetheless, we observe that our method is capable of partially addressing the class bias problem by suppressing the other classes through regularization. 
}
\begin{table}[t]
\centering
\normalsize
  \caption{Performance of weakly supervised semantic segmentation on \rv{MS COCO 2014} validation images. Sal. denotes the methods using external salient object mask supervision.}\label{table_semantic_coco}
%   Level of supervision: $\mathcal{F}-$full, $\mathcal{I}-$image label, $\mathcal{B}-$box, $\mathcal{S}-$saliency.}  
\vspace{-0.7em}
\begin{tabular}{l@{\hskip 0.2in}c@{\hskip 0.1in}cc}
    \Xhline{1pt}\\[-0.95em]
    Method  & Backbone& Sal. & mIoU \\
    \hline\hline 
    \\[-0.9em]
    
    $\text{DSRG}_{\text{~~CVPR '18}}$~\cite{huang2018weakly} & VGG16  & \checkmark  &   26.0 \\
    $\text{ADL}_{\text{~~TPAMI '20}}$~\cite{choe2020attention} & VGG16  & \checkmark  &   30.8 \\
    $\text{Yao \textit{et al.}}_{\text{~~ACCESS '20}}$~\cite{choe2020attention} & VGG16  & \checkmark  &   33.6 \\
    $\text{CONTA}_{\text{~~NeurIPS '20}}$~\cite{zhang2020causal}   & ResNet50  &   & 33.4  \\
    $\text{IRN}_{\text{~~CVPR '19}}$~\cite{ahn2019weakly} & ResNet101 &   & 41.4 \\

    AdvCAM (Ours) & ResNet101 &  & 44.4  \\
    % \\[-0.9em]
    \Xhline{1pt}
    
    \end{tabular}%
    
    % \end{adjustbox}
    \vspace{-1em}

      \end{table}

\subsection{Semi-Supervised Semantic Segmentation}
Table~\ref{tabsemi} compares the mIoU scores achieved by our method on the \rv{PASCAL VOC 2012} validation and test images with those from other recent semi-supervised segmentation methods.
All \rvt{of} these methods were implemented on the ResNet-based backbone~\cite{he2016deep}, except for the first four, which used the VGG-based backbone~\cite{simonyan2014very}.
Our method achieved mIoU scores of 77.8 and 76.9 on the PASCAL VOC 2012 validation and test images respectively, which are higher than those of the other methods under the same level of supervision. \rv{These methods include PseudoSeg~\cite{zou2020pseudoseg} and Lai \textit{et al.}~\cite{Lai2021semi}, which are contemporary with our method.}
In particular, the performance of our method on the validation images was 4.6\%p better than that of CCT~\cite{ouali2020semi}, which is our baseline. 
Our method even outperformed the method of Song \textit{et al.}~\cite{song2019box}, which uses bounding box labels, which are stronger annotations than class labels.
Fig.~\ref{segsample} compares examples of semantic masks produced by CCT~\cite{ouali2020semi} and by our method, \rv{which shows that our method captures the regions occupied by the target object more accurately than CCT~\cite{ouali2020semi}.}

\section{Experiments on Object Localization}
\subsection{Experimental Setup}
\textbf{Datasets:} For weakly supervised object localization, we used the CUB-200-2011~\cite{wah2011caltech} and ImageNet-1K~\cite{deng2009imagenet} datasets.
The CUB-200-2011 dataset contains 6,033 training images and 5,755 test images.
\rv{These images depict birds of 200 species, and each species is a class.}
The ImageNet-1K dataset contains about 1.3M training images and 50,000 test images, \rv{depicting} 1000 classes of everyday objects. For both datasets, we determine the model and hyper-parameters with validation images provided by Choe \textit{et al.}~\cite{choe2020evaluating}, and report Top-1 classification accuracy, Top-1 localization accuracy, and \texttt{MaxBoxAccV2}~\cite{choe2020evaluation} for a range of IoU thresholds on test images.
Note that \texttt{MaxBoxAccV2} at an IoU threshold of 0.5 is equivalent to the GT-known localization accuracy.

\textbf{Reproducibility:} We set the value of $\lambda$ to 0.01 for both datasets.
We used the ResNet-50~\cite{he2016deep} \rv{and Inception-V3~\cite{szegedy2016rethinking} backbone networks} pre-trained on the ImageNet dataset.

\begin{table}[tbp]
\normalsize
  \centering  \caption{Comparison of mIoU scores for semi-supervised semantic segmentation methods on the PASCAL VOC 2012 \textit{val} and \textit{test} images.}
%   \resizebox{0.48\textwidth}{!}{
\vspace{-0.7em}
\begin{threeparttable}
    \begin{tabular}{l@{\hskip 0.1in}c@{\hskip 0.2in}cc}
    \Xhline{1pt}\\[-0.95em]
    Method & Training set & \textit{val} & \textit{test}    \\
    \hline\hline\\[-0.95em]
    % DeepLab~\cite{chen2014semantic} & 1.4K strong & 62.5\\
    % \hline\\[-0.95em]
     $\text{WSSL$^{\dagger}$}_{\text{~~ICCV '15}}$~\cite{papandreou2015weakly}  & 1.5K $\mathcal{F}$ + 9.1K $\mathcal{I}$ & 64.6 & 66.2 \\
    % GAIN~\cite{li2018tell}  & 1.5K $\mathcal{F}$ + 9.1K $\mathcal{I}$ & 60.5 & 62.1 \\
    $\text{MDC$^{\dagger}$}_{\text{~~CVPR '18}}$~\cite{wei2018revisiting}  & 1.5K $\mathcal{F}$ + 9.1K $\mathcal{I}$ & 65.7 & 67.6\\
    
    % DSRG ~\cite{huang2018weakly} (baseline) & 1.4K strong + 9.1K weak& 64.3\\
    $\text{Souly \textit{et al.}$^{\dagger}$}_{\text{~~ICCV '17}}$~\cite{souly2017semi} & 1.5K $\mathcal{F}$ + 9.1K $\mathcal{I}$ &   65.8  & - \\
     $\text{FickleNet$^{\dagger}$}_{\text{~~CVPR '19}}$~\cite{lee2019ficklenet} & 1.5K $\mathcal{F}$ + 9.1K $\mathcal{I}$ &   65.8  & - \\
    $\text{Song \textit{et al.}}_{\text{~~CVPR '19}}$~\cite{song2019box}& 1.5K $\mathcal{F}$ + 9.1K $\mathcal{B}$ &   71.6 & -  \\
    $\text{CCT}_{\text{~~CVPR '20}}$~\cite{ouali2020semi} & 1.5K $\mathcal{F}$ + 9.1K $\mathcal{I}$  &   73.2 & -  \\
    $\text{PseudoSeg}_{\text{~~ICLR '21}}$~\cite{zou2020pseudoseg}& 1.5K $\mathcal{F}$ + 9.1K $\mathcal{I}$ &   73.8 & -  \\
    $\text{Lai \textit{et al.}}_{\text{~~CVPR '21}}$~\cite{Lai2021semi} & 1.5K $\mathcal{F}$ + 9.1K $\mathcal{I}$ &   76.1 & -  \\

    $\text{Luo \textit{et al.}}_{\text{~~ECCV '20}}$~\cite{luosemi}& 1.5K $\mathcal{F}$ + 9.1K $\mathcal{I}$ &   76.6 & -  \\
    
    AdvCAM (Ours) & 1.5K $\mathcal{F}$ + 9.1K $\mathcal{I}$ &   \textbf{77.8}  &  \textbf{76.9}\\
    % advCAM (Ours) & 1.5K $\mathcal{F}$ + 9.1K $\mathcal{I}$ &   \textbf{76.9}  &  \\

    % \hline \\[-0.95em]
    % DeepLab~\cite{chen2014semantic} & 10.6K strong & 67.6  \\
    \Xhline{1pt}
    \end{tabular}%
    \begin{tablenotes}
  \footnotesize
\item $\mathcal{F}-$\rv{full}, $\mathcal{I}-$\rv{image class}, $\mathcal{B}-$box, $^{\dagger}-$ VGG backbone \\
% $^*$ Anonymous result link can be found \href{http://host.robots.ox.ac.uk:8080/anonymous/TQSZTQ.html}{here}.
\scriptsize
% $^*$\url{http://host.robots.ox.ac.uk:8080/anonymous/TQSZTQ.html}
        \end{tablenotes}
     \end{threeparttable}
  \label{tabsemi}
%   }%
    % \vspace*{-0.55\baselineskip}
    \vspace{-1.3em}

\end{table}%

\begin{table*}[htbp]
\normalsize
\renewcommand{\arraystretch}{1.1}
  \centering
  \caption{Weakly supervised object localization performance on CUB-200-2011 and ImageNet-1K test images. All these results were obtained using a ResNet-50 backbone.}
  \vspace{-1em}
    \begin{tabular}{l|c@{\hskip 0.1in}ccccc|c@{\hskip 0.1in}ccccc}
    \Xhline{1pt}
          & \multicolumn{6}{c}{CUB-200-2011}      & \multicolumn{6}{@{\hskip -0.005in}|c}{ImageNet-1K} \\
          \hline
    \multicolumn{1}{l|}{\multirow{2}[0]{*}{Method}} & Top-1 & Top-1  & \multicolumn{4}{c|}{\texttt{MaxBocAccV2}@IoU(\%)} & Top-1  & Top-1 & \multicolumn{4}{c}{\texttt{MaxBocAccV2}@IoU(\%)} \\
    \multicolumn{1}{c|}{} & Cls (\%) & Loc (\%) & 0.3   & 0.5   & 0.7  & Mean & Cls (\%) & Loc (\%) & 0.3   & 0.5   & 0.7 & Mean\\
    \hline\hline 
    
    $\text{HaS}_{\text{~~ICCV '17}}$~\cite{singh2017hide}   & 69.7      &   -    &     93.1  &  72.2      &    28.6  &  64.6 &    75.4   &    -   &   83.7    &     63.2  & 41.3 & 62.7 \\
    $\text{ACoL}_{\text{~~CVPR '18}}$~\cite{zhang2018adversarial}   &    71.1   &   -    & 97.0       &    77.3   &   25.0  &  66.4 &   73.1    &   -    &   83.5    &  64.4     & 39.0  & 62.3 \\
    $\text{SPG}_{\text{~~ECCV '18}}$~\cite{zhang2018self}   &     50.5  &    -   &  92.2     &     68.2  &      20.8 & 60.4   &  73.3  &  -     &    83.9   &      65.4  & 40.6 & 63.3\\
    $\text{ADL}_{\text{~~TPAMI '20}}$~\cite{choe2020attention}   & 66.6      &  -     &     91.8  &     64.8  &      18.4 & 58.3   & 71.2      &    -   &   83.6    &     65.6  & 41.8 & 63.7 \\
    $\text{Ki \textit{et al.}}_{\text{~~ACCV '20}}$~\cite{ki2020sample}   &   \textbf{80.4}    &  56.1     &    96.2   &      72.8  &     20.6  &     63.2 &   71.3    &  48.4     &  84.3     &  67.6     &  43.6 & 65.2\\
    % $\text{MEIL}_{\text{~~CVPR '20}}$~\cite{mai2020erasing}   &      &     &       &       &     &    &      &    &     &      &   & \\
    $\text{Bae \textit{et al.}}_{\text{~~ECCV '20}}$~\cite{bae2020rethinking}   &    75.0   &  59.5     &     -  &    77.6  &    -  & - &  75.8  &  49.4    &  -   &  62.2     &    -   & -  \\
    $\text{I$^2$C}_{\text{~~ECCV '20}}$~\cite{zhang2020inter}   &    -   &  56.0     &     -  &    72.6   &    -   &  -  &  76.7    &  54.8     &  -     &    68.5   & - & -  \\
    $\text{DGL}_{\text{~~ACMMM '20}}$~\cite{tan2020dual}   & -  &      60.8 &     -  &   74.7    & -      &   -   & -  &  53.4     &  -     &    66.5   & - & - \\
    % $\text{GC-Net}_{\text{~~ECCV '20}}$~\cite{lu2020geometry}   &    76.8   &      63.2 &       &       &       &      &  77.4 &  49.1     &  -     &    -   & - & - \\
    $\text{Babar \textit{et al.}}_{\text{~~WACV '21}}$~~\cite{babar2021look}   &    77.3   &      64.7 &     -  &    77.4   &    -   &   -   &  - &  52.4     &  -     & 67.9   & - & - \\
    
    \hline
    $\text{CAM}_{\text{~~CVPR '16}}$~\cite{zhou2016learning}   &    77.6   &  58.0   &   95.3  &    71.5  &  19.0 &  61.9  &    75.5  &  51.9   &    83.9  &     65.8 & 41.5  & 63.7\\
    CAM + Ours   &   77.6   & 63.7     &  \textbf{97.8}   &  80.5     &   27.4    & 68.6   &  75.5   &   \textbf{55.0}   &    \textbf{86.5} &  \textbf{70.3}    &  \textbf{47.4}  & \textbf{68.1}\\
    $\text{CutMix}_{\text{~~ICCV '19}}$~\cite{yun2019cutmix}   &  78.3    &  57.6  &  93.1     &  70.8     &   23.4 &  62.4  &  \textbf{77.9}     &  52.8     &  84.6     & 65.4     &  41.6 & 63.9 \\
    CutMix + Ours   &  78.3     &   \textbf{64.9}    &    97.3  &    \textbf{80.7}   &    \textbf{32.5}   &   \textbf{70.2} & \textbf{77.9} & 54.8  &    86.2  & 68.4   &     45.5  & 66.7  \\

    \Xhline{1pt}
    \end{tabular}%
  \label{table_wsol}%
\end{table*}%

\begin{figure*}[t]
\centering
\vspace{-0.2em}
\includegraphics[width=\linewidth]{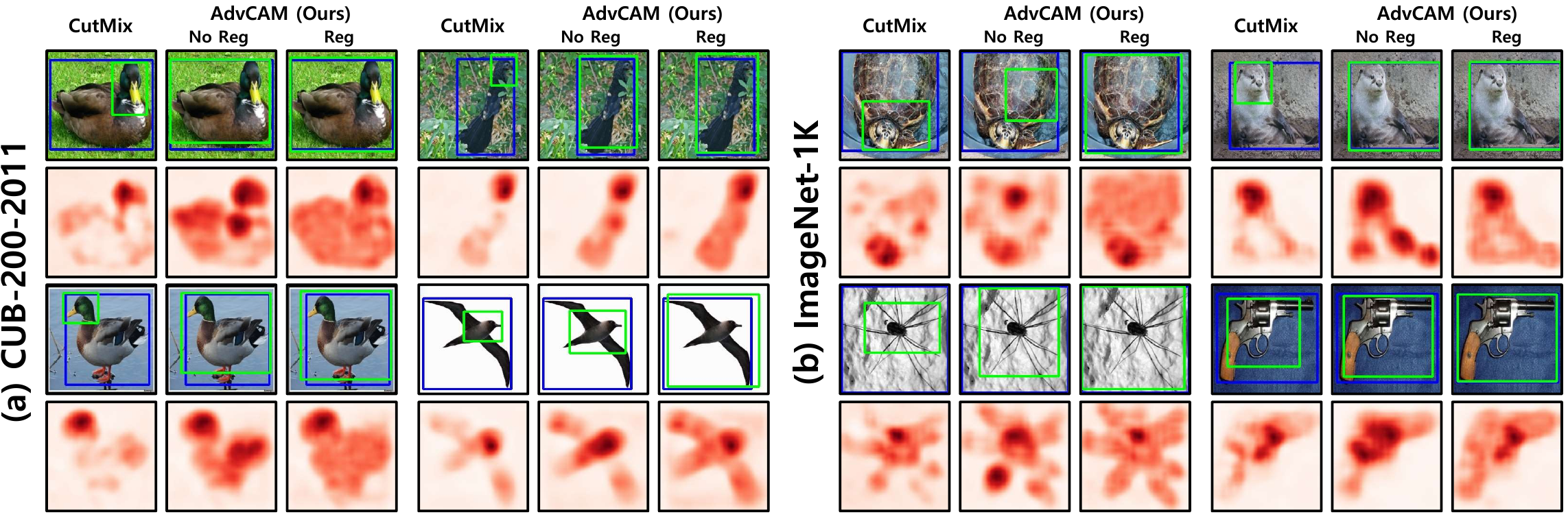}
\vspace{-1.5em}
\caption{\label{wsol_sample_cub_imagenet} Examples of images from the (a) CUB-200-2011 and (b) ImageNet-1k datasets with predicted bounding boxes and localization maps ($\mathcal{A}$) from CutMix~\cite{yun2019cutmix} and our method. The blue \rv{boxes represent} the ground truth, and the green \rv{boxes represent} the predictions.}
\vspace{-1em}
\end{figure*}

\subsection{Experimental Results}

\textbf{Results on CUB-200-2011.} 
Table~\ref{table_wsol} compares results \rv{from} our method and \rv{from} existing methods on the CUB-200-2011~\cite{wah2011caltech} dataset, \rv{using a ResNet-50~\cite{he2016deep} backbone.} Our method achieves a 7.3\%p improvement in Top-1 localization accuracy over CutMix~\cite{yun2019cutmix}, which is our baseline, \rvt{while maintaining the} same classification accuracy. In addition, the \texttt{MaxBoxAccV2} scores for IoU thresholds of 0.5 and 0.7 increase by 9.9\%p and 9.1\%p respectively, indicating that adversarial climbing produces more exact bounding boxes.
Table~\ref{table_wsol} shows that our method also yields significantly better results than those \rv{from} other recently published methods.
Since adversarial climbing can be performed on any differentiable model, we expect our method to be applicable to methods other than CAM and CutMix without sacrificing classification accuracy.

\rv{In addition, we compare \texttt{MaxBoxAccV2}~\cite{choe2020evaluation} scores obtained by the above methods using the Inception-V3~\cite{szegedy2016rethinking} backbone.
Table~\ref{table_inception_wsol} shows that this improves the \texttt{MaxBoxAccV2} scores for IoU thresholds of 0.5 and 0.7 by over 10\%p, compared to our CutMix baseline~\cite{yun2019cutmix}.
Table~\ref{table_inception_wsol} also shows that our method outperforms even the most recent methods by a large margin. In particular, the \texttt{MaxBoxAccV2} score is 3.7\%p higher than that of IVR~\cite{kim2021normalization}, the best-performing method among our competitors.
}

\textbf{Results on ImageNet-1K.} Table~\ref{table_wsol} compares results \rv{from} our method and \rv{from} existing methods on the ImageNet-1K dataset~\cite{deng2009imagenet}, \rv{using a ResNet-50~\cite{he2016deep} backbone.} Our method achieved a new state-of-the-art performance on the ImageNet-1K dataset, with a 3.1\%p increase in Top-1 localization accuracy and a 4.4\%p increase in \texttt{MaxBoxAccV2}, \rv{compared to our CAM baseline~\cite{zhou2016learning}.}

Fig.~\ref{wsol_sample_cub_imagenet} shows examples of localization maps and bounding boxes generated by CutMix~\cite{yun2019cutmix} and by our method, on the CUB-200-2011 and ImageNet-1K datasets. 
We see that the more accurate masks provided by adversarial climbing lead to more accurate bounding boxes.

\begin{figure*}[t]
\centering
\includegraphics[width=\linewidth]{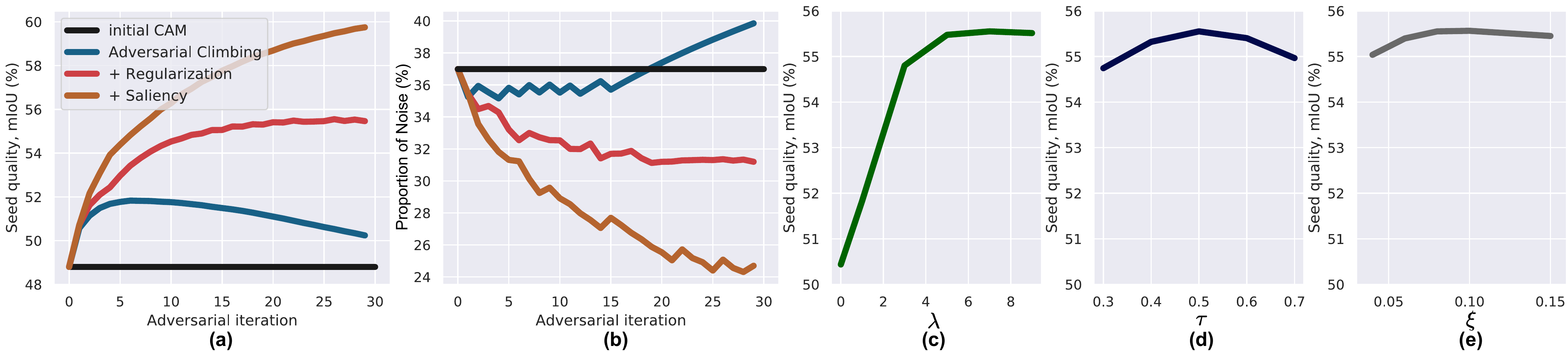}
\vspace{-2em}
\caption{\label{eachiter} Ablation studies on \rv{PASCAL VOC 2012}.
Effect of adversarial climbing and regularization on (a) seed quality and (b) the proportion of noise. (c) Effect of the regularization coefficient $\lambda$. (d) Effect of the masking threshold $\tau$. (d) Effect of the step size $\xi$.}
% Effectiveness of regularization technique. (\textit{Left}) Seed quality on mIoU (\%). (\textit{Right}) 1-precision (\%)}
\vspace{-.5em}
\end{figure*}

\begin{figure*}[t]
\centering
\vspace{-0.2em}
\includegraphics[width=\linewidth]{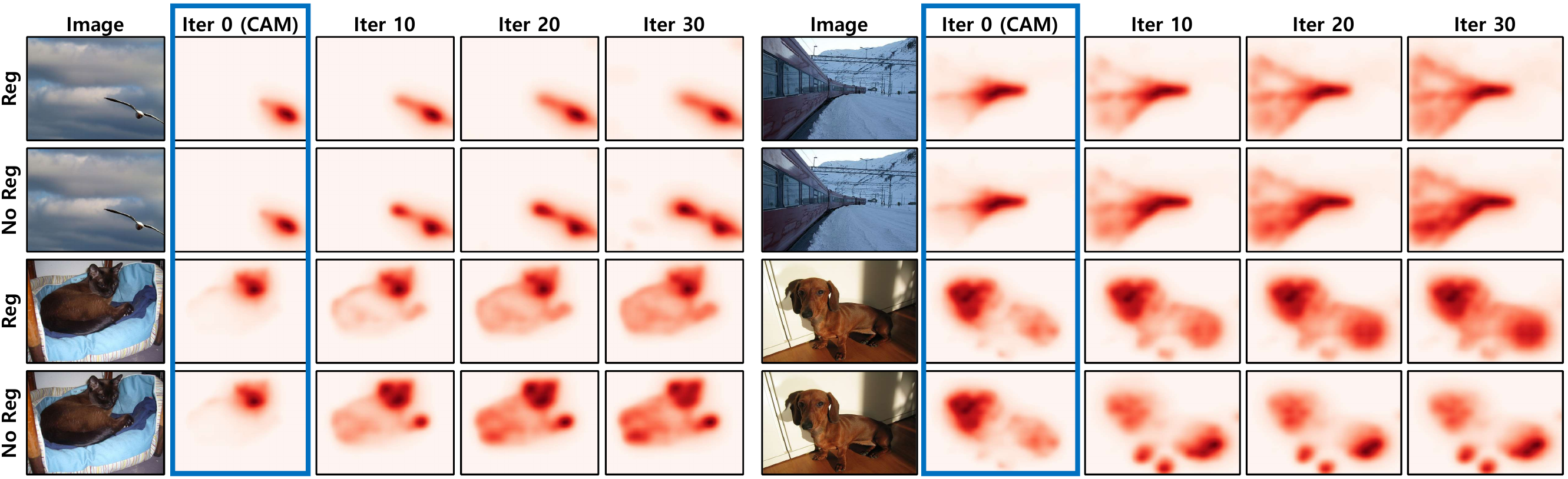}
\vspace{-2em}
\caption{\label{adv_climb_ex_voc} Four examples of initial CAMs (the blue boxes) and successive localization maps (\rv{$\mathcal{A}$}) obtained from images manipulated by iterative adversarial climbing, with the regularization procedure (top row in each example) and without (bottom row).}
% Effectiveness of regularization technique. (\textit{Left}) Seed quality on mIoU (\%). (\textit{Right}) 1-precision (\%)}
% \vspace{-1.2em}
\vspace{-1.em}
\end{figure*}

\section{Discussion}

\subsection{Iterative Adversarial Climbing}\label{iterative}

We analyzed the effectiveness of the iterative adversarial climbing and regularization technique introduced in Section~\ref{reg_sec} by evaluating the initial seed for each adversarial iteration using images from the \rv{PASCAL VOC 2012} dataset.
Fig.~\ref{eachiter}(a) shows the resulting mIoU scores: they rise steeply in the beginning, with or without regularization, but without regularization, the curves peak around iteration 8.

\rv{To take a closer look at this observation}, we evaluated the quality of the newly localized region at each \rv{iteration of} adversarial climbing in terms of the proportion of noise, which we define to be the proportion of pixels that \rvt{are classified as the foreground but actually belong in the background.} Without regularization, the proportion of noise rises steeply after around 15 iterations, as shown in Fig.~\ref{eachiter}(b): 
\rv{this implies that the new regions identified in subsequent iterations belong mostly in \rvt{the} background.}
Regularization \rv{brings the proportion of noise below that in the original CAM,}
indicating that new regions of the target object are \rv{still} being found \rv{during} as many as 30 adversarial \rvt{iterations}.
We also see that employing a salient object detector during adversarial climbing provides a better initial seed in terms of mIoU and \rvt{the} proportion of noise.
Fig.~\ref{adv_climb_ex_voc} shows \rv{examples of localization maps} at each adversarial iteration with and without regularization.
\rv{As adversarial climbing proceeds, the localization maps gradually cover more of the target object, whether there is regularization or not.}
\rv{However, without regularization, some regions of a target object (\textit{e.g.,} the legs of a dog) \rvt{may} have greatly raised attribution scores, and thus the other regions corresponding to the target object (\textit{e.g., }the dog's head) are suppressed by \rvt{the} normalization effect. Regularization allows the target object to be activated evenly, without over-activation on specific parts.}

We will now look at the effectiveness of adversarial climbing and regularization in weakly supervised object localization on the CUB-200-2011 dataset.
We obtained localization results from CutMix~\cite{yun2019cutmix}\rv{, and then} applied adversarial climbing and the \rv{proposed regularization technique}.
Table~\ref{ablation_wsol} shows that adversarial climbing improved both the Top-1 localization accuracy and the \texttt{MaxBoxAccV2} scores, and that regularization \rvt{further improved both of} them.
\rv{The improvement to the} \texttt{MaxBoxAccV2} scores is particularly noticeable at high IoU thresholds, \rv{showing that the bounding boxes produced by our method are closer approximations to the ground-truth boxes.}
As we have already noted, our method is a post-hoc analysis of a trained classifier, so its classification results are not affected.

\subsection{Hyper-Parameter Analysis}\label{hyperparam}
Having looked at the effect of the number of adversarial iterations (Figs.~\ref{eachiter}(a) and (b)), 
we now analyze the sensitivity of the mIoU of the initial seed to the other three hyper-parameters involved in adversarial climbing. 
All the results reported in this section were obtained without using a salient object detector to focus on the sensitivity of adversarial climbing to each hyper-parameter.

\begin{table}[t]
  \centering
  \normalsize
\color{black}

  \caption{\rv{Comparison of \texttt{MaxBoxAccV2} scores on the CUB-200-2011 dataset using the Inception-V3~\cite{szegedy2016rethinking} backbone.}}
  \vspace{-0.5em}
    \begin{tabular}{lcccc}
    \Xhline{1pt}\\[-0.95em]
    \multicolumn{1}{l}{\multirow{2}[0]{*}{Method}} & \multicolumn{4}{c}{\texttt{MaxBocAccV2}@IoU(\%)} \\
    \multicolumn{1}{c}{} & 0.3   & 0.5   & 0.7   & Mean \\
        \hline\hline 

    % DANet$_{\text{~~CVPR '19}}$~\cite{xue2019danet} &  & 49.5  &   - \\
    
    ADL$_{\text{~~CVPR '19}}$~\cite{choe2020attention} & 93.8  & 65.8  &  16.9 & 58.8 \\
    DGL$_{\text{~~ACMMM '20}}$~\cite{tan2020dual} &  - &  67.6 & - & -    \\
    Ki \textit{et al.}$_{\text{~~ACCV '20}}$~\cite{ki2020sample} & \textbf{95.9}  & 67.9  & 17.2 &  60.3 \\
    Bae \textit{et al.}$_{\text{~~ECCV '20}}$~\cite{bae2020rethinking} &  - &  70.0 & - & -    \\
    CALM$_{\text{~~ICCV '21}}$~\cite{kim2021keep} & -  &  - &  - &60.3 \\

    IVR$_{\text{~~ICCV '21}}$~\cite{kim2021normalization} & -  &  - &  - &61.7 \\
    \\[-0.95em]
    \hline
    \\[-0.95em]
    CutMix$_{\text{~~ICCV '19}}$~\cite{yun2019cutmix} & 91.4  & 62.2  &  18.0 & 57.2 \\
    CutMix + Ours & 95.5  & \textbf{72.4}  & \textbf{28.2} & \textbf{65.4}  \\
    \Xhline{1pt}
    \end{tabular}%
    \vspace{-1em}
  \label{table_inception_wsol}%
\end{table}%

\textbf{Regularization Coefficient $\boldsymbol{\lambda}$:} Eq.~\ref{reg_loss} shows how $\lambda$ controls the strength of the regularization that limits the extent to which adversarial climbing can increase the attribution scores of regions that already have high scores.
Fig.~\ref{eachiter}(c) shows the mIoU of the initial seed for different values of $\lambda$. 
Regularization improves the quality of initial seeds by more than 5\%p (50.43 for $\lambda=0$, when there is no regularization, \textit{vs.} 55.55 for $\lambda=7$).
The curve \rvt{plateaus} after $\lambda=5$, suggesting that it is not difficult to select a good value of $\lambda$.

\textbf{Masking Threshold $\boldsymbol{\tau}$}: 
By controlling the size of the restricting mask $\mathcal{M}$ in Eq.~\ref{mask}, \rv{the hyper-parameter} $\tau$ determines how many pixels \rv{will retain} attribution values \rvt{similar} to those of the original CAM during adversarial climbing.
Fig.~\ref{eachiter}(d) shows the mIoU of the initial seed for different values of $\tau$. 
We can see that $\tau$ influences the quality of the initial seeds less than $\lambda$: varying $\tau$ from 0.3 to 0.7 produces a change of less than 1\%p in mIoU. Again, it should be straightforward to select a good value of $\tau$.
\begin{table*}[t]
  \centering
  
\begin{minipage}{0.56\linewidth}
\normalsize
  \centering
  \includegraphics[width=1\linewidth]{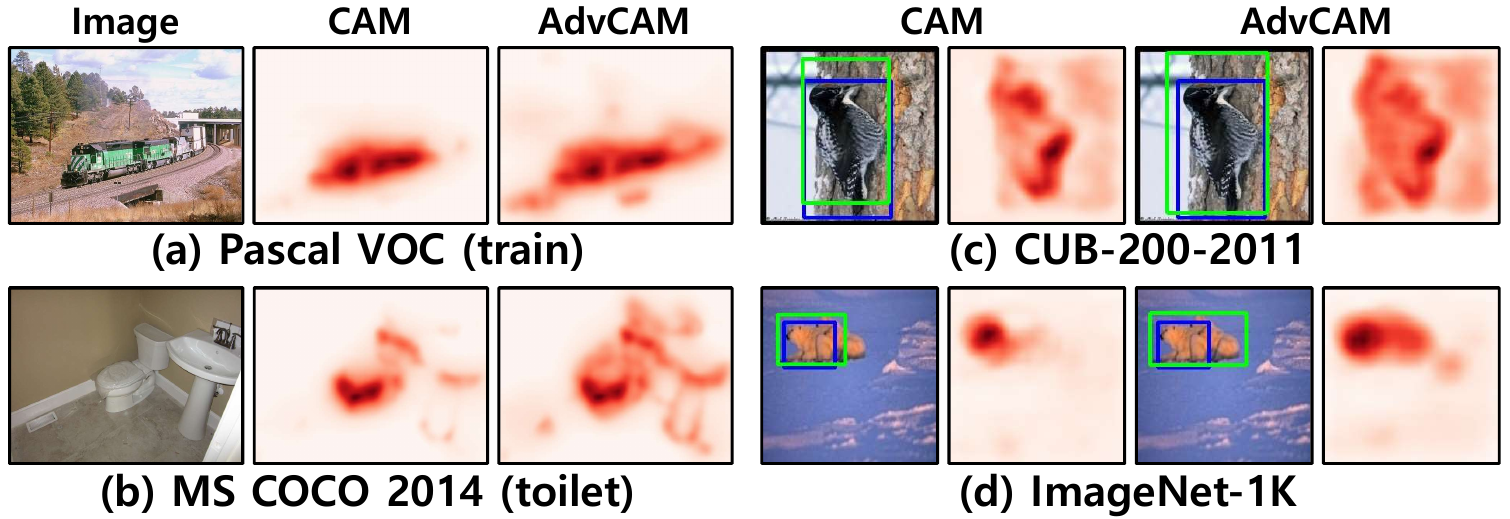}
  \vspace{-2em}
  \captionof{figure}{\label{fig_failure} \normalsize \rv{Failure examples for each dataset. For (c) and (d), the blue box represents the ground truth, and the green box represents the prediction. }}

  \end{minipage}\hfill
%   Table2
\begin{minipage}{0.42\linewidth}
\center
{\normalsize
\vspace{1em}
\color{black}
\begin{tabular}{l@{\hskip 0.in}c@{\hskip 0.1in}c@{\hskip 0.1in}c}
    \Xhline{1pt}\\[-0.95em]
    Method  & Precision & Recall & F1-score \\
    \hline\hline 
    \\[-0.9em]
    IRN$_{\text{~~CVPR '19}}$~\cite{ahn2019weakly} & 66.0  & 66.4  &   66.2 \\
    Chang \textit{et al.}$_{\text{~~CVPR '20}}$~\cite{chang2020weakly} &  61.0 & 77.2  &   68.1 \\
    SEAM$_{\text{~~CVPR '20}}$~\cite{wang2020self} & 66.8  & 76.8  &   71.5 \\
    AdvCAM (Ours)   & 66.9  & 77.6  & 71.9  \\
    AdvCAM--Sal (Ours) & \textbf{71.8} &  \textbf{78.2} & \textbf{74.9} \\

    % \\[-0.9em]
    \Xhline{1pt}
    \end{tabular}%
    }
    \vspace{0.2em}
      \caption{\rv{Comparison of precision, recall, and F1-score on the \rv{PASCAL VOC 2012} \rvt{\textit{train}} images.}}
      \label{table_precrecall}%

  \end{minipage}\hfill
\vspace{-1em}
\end{table*}

\textbf{Step Size $\boldsymbol{\xi}$}: This determines the extent of the adversarial manipulation of the image in Eq.~\ref{sgd_reg}. Fig.~\ref{eachiter}(e) shows the mIoU of the initial seed for different values of $\xi$. A broad range of step sizes are satisfactory.

\begin{table}[t]
\normalsize
\renewcommand{\arraystretch}{1.1}
  \centering
  \caption{Effects of adversarial climbing and regularization on the performance of weakly supervised object localization for the CUB-200-2011 dataset, using CutMix~\cite{yun2019cutmix}  based on ResNet-50.}
    \begin{tabular}{cc|ccccc}
    \Xhline{1pt}
          
    Adv.& \multicolumn{1}{c|}{\multirow{2}[0]{*}{Reg.}} & Top-1  & \multicolumn{4}{c}{\texttt{MaxBocAccV2}@IoU(\%)} \\
    Climb & \multicolumn{1}{c|}{}  & Loc (\%) & 0.3   & 0.5   & 0.7  & Mean \\
    \hline\hline 
     \textcolor{red}{\xmark} & \textcolor{red}{\xmark} & 57.6 & 93.1 & 70.8 & 23.4 & 62.4 \\
     
     \textcolor{OliveGreen}{\cmark}  &  \textcolor{red}{\xmark}    &    62.5   &   \textbf{98.0} &    78.2   &    26.4   &   67.6  \\
     \textcolor{OliveGreen}{\cmark}  &  \textcolor{OliveGreen}{\cmark}   &   \textbf{64.9}    &    97.3  &    \textbf{80.7}   &    \textbf{32.5}   &   \textbf{70.2}   \\

    \Xhline{1pt}
    \end{tabular}%
    % \vspace{-1em}
  \label{ablation_wsol}%
\end{table}%

\begin{table}[t]
\normalsize
\color{black}
  \centering
  \caption{\rv{Effects of design choice of the final localization map on the quality of the initial seed.} }
%   \vspace{-1em}
%   Table A1: Performance improvement of IRN [\textcolor{green}{2}] by ours and CONTA~[\textcolor{green}{61}] for the seed and the pseudo mask (Mask) on COCO \textit{train} set, and the segmentation (Seg) on COCO \textit{val} set.}
    \begin{tabular}{lcccc}
     \Xhline{1pt}\\[-1.em]
    Design  &  mIoU  & Precision & Recall & F1-score \\ 
    \hline\hline\\[-0.9em]
    $\texttt{CAM}(x^{T})$ & 54.0 & 64.4 & \textbf{78.2} & 70.6 \\

    $\sum_{t=0}^{T} \texttt{CAM}(x^{t})$ &  \textbf{55.6} & \textbf{66.9} & 77.6 & \textbf{71.9} \\
    \\[-0.95em]
    \Xhline{1pt}
    \vspace{-2em}
    \end{tabular}%
    
  \label{design_choice_table}%
\end{table}%

\begin{table}[t]
\normalsize
  \centering
  \caption{Effects of adversarial climbing on different methods of generating the initial seed: mIoU of the initial seed (Seed) and of the pseudo ground truth mask (Mask), for the PASCAL VOC 2012 \rvt{\textit{train}} images.}
  \vspace{-0.7em}
    \begin{tabular}{l@{\hskip 0.4in}l@{\hskip 0.2in}l}
     \Xhline{1pt}\\[-0.95em]
    Method  & Seed  & Mask \\
    \hline\hline \\[-0.9em]
    $\text{Grad-CAM++}$~\cite{chattopadhay2018grad} & 47.7  & ~~-  \\
    + Adversarial climbing  &50.7$_{~+3.0}$  & ~~-   \\
    \hline \\[-0.9em]
    $\text{Chang \textit{et al.}}$~\cite{chang2020weakly} & 50.9 & 63.4 \\
    + Adversarial climbing  & 53.7$_{~+2.8}$ & 67.5$_{~+4.1}$  \\
    \hline \\[-0.9em]
    $\text{SEAM}$~\cite{wang2020self} & 55.4  & 63.6  \\
    + Adversarial climbing  & 58.6$_{~+3.2}$ & 67.2$_{~+3.6}$   \\
    \hline \\[-0.9em]
    $\text{IRN}$~\cite{ahn2018learning} & 48.8  & 66.3  \\
    + Adversarial climbing  &55.6$_{~+6.8}$  & 69.9$_{~+3.6}$   \\
    % \hline\\[-0.95em]
    % advCAM (Ours)  & \xmark  & 36.3 & 13.1 \\
    % advCAM (Ours) & \textcolor{red}{\cmark} & - & - \\
    \Xhline{1pt}
    % \vspace{-2.em}
    \end{tabular}%
    \vspace{-0.5em}

  \label{tab:baselines}%
\end{table}%

% \begin{table}[htbp]
% % \renewcommand{\arraystretch}{0.97}
%   \centering
%   \caption{Different baselines + advCAM on PASCAL VOC 2012 train images in mIoU(\%)}
%   \vspace{-0.7em}
%     \begin{tabular}{cccc}
%      \Xhline{1pt}\\[-0.95em]
%     advCAM  & Chang \textit{et al.}~\cite{chang2020weakly}  & SEAM~\cite{wang2020self} & IRN~\cite{ahn2019weakly}\\
%     \hline\hline \\[-0.9em]
%      -  & 50.9 & 63.4 \\
%     \checkmark  & 52.9$_{~\textcolor{red}{\scaleto{+2.0}{6pt}}}$ & 66.3$_{~\textcolor{red}{\scaleto{+2.9}{6pt}}}$  \\
%     \hline \\[-0.9em]
%     - & 55.4  & 63.6  \\
%     \checkmark  & 57.5$_{~\textcolor{red}{\scaleto{+2.1}{6pt}}}$ & 64.4$_{~\textcolor{red}{\scaleto{+0.8}{6pt}}}$   \\
%     \hline \\[-0.9em]
%     - & 48.8  & 66.3  \\
%     \checkmark  & 55.6$_{~\textcolor{red}{\scaleto{+6.8}{6pt}}}$  & 69.4$_{~\textcolor{red}{\scaleto{+3.1}{6pt}}}$   \\
%     % \hline\\[-0.95em]
%     % advCAM (Ours)  & \xmark  & 36.3 & 13.1 \\
%     % advCAM (Ours) & \textcolor{red}{\cmark} & - & - \\
%     \Xhline{1pt}
%     \end{tabular}%
%   \label{tab:baselines}%
% \end{table}%

\subsection{\rv{Design Choice of the Final Localization Map $\mathcal{A}$}}\label{finalmap}
\rv{We analyze the effects of the design choice of the final localization map $\mathcal{A}$. As mentioned in Section~\ref{Advcam_method}, we compute the final localization map by aggregating the CAMs produced during all the adversarial climbing iterations ($\mathcal{A} = \sum_{t=0}^{T} \texttt{CAM}(x^{t})$) to suppress the noise that occurs in the later iterations. 
Alternatively, a final localization map could be obtained just from the CAM produced during \rvt{the last} adversarial climbing iteration ($\mathcal{A} = \texttt{CAM}(x^{T})$). 
Table~\ref{design_choice_table} compares the mIoU, precision, recall, and F1-score values that result from each method of computing the final localization map.
The final localization map computed from $\texttt{CAM}(x^{T})$ alone achieves high recall but very low precision, indicating that a lot of \rvt{the} background is receiving a high attribution score. 
Aggregating CAMs over adversarial climbing iterations ($\mathcal{A} = \sum_{t=0}^{T} \texttt{CAM}(x^{t})$) achieves much higher precision while maintaining recall; this demonstrates that the aggregation process largely prevents background regions from being identified.
}

\subsection{Generality of Our Method}\label{generality}
Most of our experiments with adversarial climbing were based on IRN~\cite{ahn2019weakly}.
However, we also applied adversarial climbing to two state-of-the-art methods of generating an initial seed for weakly supervised semantic segmentation: that \rvt{of} Chang \textit{et al.}~\cite{chang2020weakly}, and SEAM~\cite{wang2020self}.
We used \rv{the} pre-trained classifiers provided by Chang \textit{et al.} and the authors of SEAM.
However, we had to train IRN's classifier since the authors~\cite{ahn2019weakly} do not provide a pre-trained one. 
We also followed the authors' experimental settings, including their choice of \rvt{the} backbone network and \rvt{the} mask refinement method. Thus, we used PSA~\cite{ahn2018learning} to refine the initial seed \rv{obtained} from the method of Chang \textit{et al.} and from SEAM, when these were combined with adversarial climbing.
Table~\ref{tab:baselines} \rvt{provides} mIoU values for the initial seed and the pseudo ground truth mask obtained when each method was combined with adversarial climbing, which was found to improve the quality of the initial seed by more than 4\%p on average. 
Our approach does not require initial seed generators to be modified or retrained.

We also thought that it would be appropriate to assess the effect of adversarial climbing on attribution methods other than the CAM~\cite{zhou2016learning}. We experimented with Grad-CAM++~\cite{chattopadhay2018grad} because Grad-CAM~\cite{selvaraju2017grad} is essentially \rv{equivalent to} CAM for ResNet~\cite{he2016deep}.
Adversarial climbing improved the results from Grad-CAM++ by 3\%p, as shown in Table~\ref{tab:baselines}.

\begin{figure*}[t]
\centering
\includegraphics[width=\linewidth]{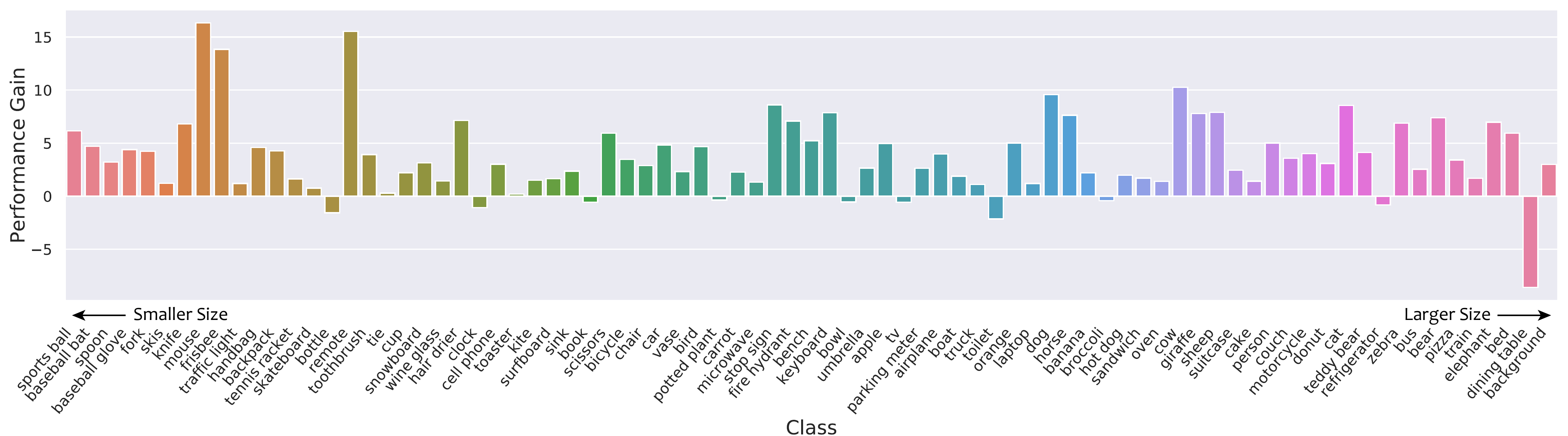}
\vspace{-2em}
\caption{\label{perclassgain} Per-class seed quality improvement in mIoU (\%p) of IRN~\cite{ahn2019weakly} \rv{achieved} by adversarial climbing on \rv{the MS COCO 2014} training images. Classes are sorted by their average sizes. \rv{These sizes are due to Choe \textit{et al.}~\cite{choe2020attention}.}}
% \vspace{-1em}
\end{figure*}

\begin{table*}[htbp]
  \centering
  \normalsize
    \color{black}
  \caption{\rv{Precision, recall, and F1-score averaged across all classes, the smallest 10 classes, and the largest 10 classes. All the results are computed for 6\% of the \rvt{images} from the MS COCO 2014 dataset ($\approx$ 5000 images). The average sizes for each class were borrowed from the work of Choe \textit{et al.}~\cite{choe2020attention}.}}
  \vspace{-0.8em}
    \begin{tabular}{l|ccc|ccc|ccc}
    \Xhline{1pt}
          & \multicolumn{3}{c|}{All Classes} & \multicolumn{3}{c|}{Smallest 10 classes} & \multicolumn{3}{c}{Largest 10 classes} \\
        %   \\[-0.95em]\cline{2-10}\\[-0.95em]
          & Precision & Recall & F1-score & Precision & Recall & F1-score & Precision & Recall & F1-score \\
          \hline\hline
    CAM   &   44.5    &    61.6   &    47.6   & 11.5      &  \textbf{73.0}     &    19.2   &    69.3   &     60.4  & 63.5 \\
    AdvCAM ($T$=10, $\tau$=0.5) &   46.7    &   63.8    &     50.6  &  14.8     &     69.7  &     23.8  &    70.9   &    63.9   & 65.6 \\
    AdvCAM ($T$=20, $\tau$=0.5) &   47.1    &   \textbf{64.7}    &    51.3   &   16.5    &   67.4    &   25.4    &  70.9     &  66.0     & 66.5 \\
    AdvCAM ($T$=30, $\tau$=0.5) &    \textbf{48.1}   &    63.6   &    \textbf{51.6}   &  18.5    &  64.9     &   27.2   &   71.4    &    65.7   &  \textbf{66.7}\\
    AdvCAM ($T$=30, $\tau$=0.4) &  48.0     & 63.6      &   51.4    &     17.1  & 66.6      & 26.1      &    \textbf{71.5}   &  64.9     & 66.3 \\
    AdvCAM ($T$=30, $\tau$=0.6) &    47.7   &    63.7   &   51.4    &   \textbf{19.2}    &   64.3    &   \textbf{28.1}    & 71.2      &   \textbf{66.1}    & 66.5 \\
    \Xhline{1pt}
    \end{tabular}%
        \vspace{-1em}

  \label{tab_size_pr}%
\end{table*}%

\subsection{\rv{Analysis of Failure Cases}}
\rv{In this section, we analyze the cases where our method did not work properly.
Fig.~\ref{fig_failure} presents some common examples of failures \rvt{within} each dataset.
Objects that are not part of the target class but are related to that class (\textit{e.g.,} train and rail in Fig.~\ref{fig_failure}(a) or toilet and sink in Fig.~\ref{fig_failure}(b)) can be also activated by adversarial climbing, which will reduce precision.
This is a long-standing problem that commonly occurs in other recent methods as well.
To analyze this quantitatively, we compare the precision, recall, and F1-score of our method with those of the other methods in Table~\ref{table_precrecall}.
Chang \textit{et al.}~\cite{chang2020weakly} achieve high recall, at the cost of a large drop in precision.
SEAM~\cite{wang2020self} avoids this loss of precision through pixel-level refinement using an additional module mentioned in Section~\ref{initialseedexp}.
Our method achieves better recall and precision without an external module.
AdvCAM--Sal further improves precision with the help of auxiliary salient object mask supervision.

We also provide some examples of failures in weakly supervised object localization in Figs.~\ref{fig_failure}(c) and (d).
Again, parts of the background are sometimes activated together with the foreground (Fig.~\ref{fig_failure}(c)).
When several objects of the target class appear in an image from the ImageNet-1K dataset, only one of them is labeled with a bounding box (see the blue box in Fig.~\ref{fig_failure}(d)). If adversarial climbing appropriately identifies all the target class regions in an image, then the IoU between the predicted box and the box label can actually be reduced, even though AdvCAM has appropriately identified all the target class regions.
}

\begin{figure*}[t]
\centering
\includegraphics[width=\linewidth]{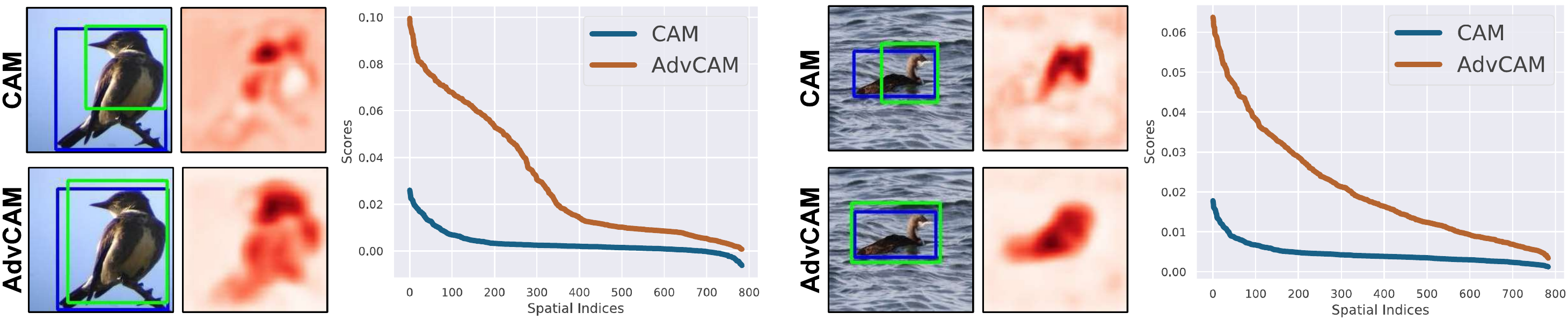}
\vspace{-1.8em}
\caption{\label{fig_bg_suppress} \rv{Two examples showing (left-hand images) the suppression of (normalized) attribution scores in background regions of an image; and (right-hand images) the sorted distribution of attribution scores of each pixel in the corresponding CAM and AdvCAM. The blue boxes represent ground truth, and the green boxes represent predictions.}}
\vspace{-1em}
\end{figure*}

\begin{figure}[t]
\centering
\includegraphics[width=\linewidth]{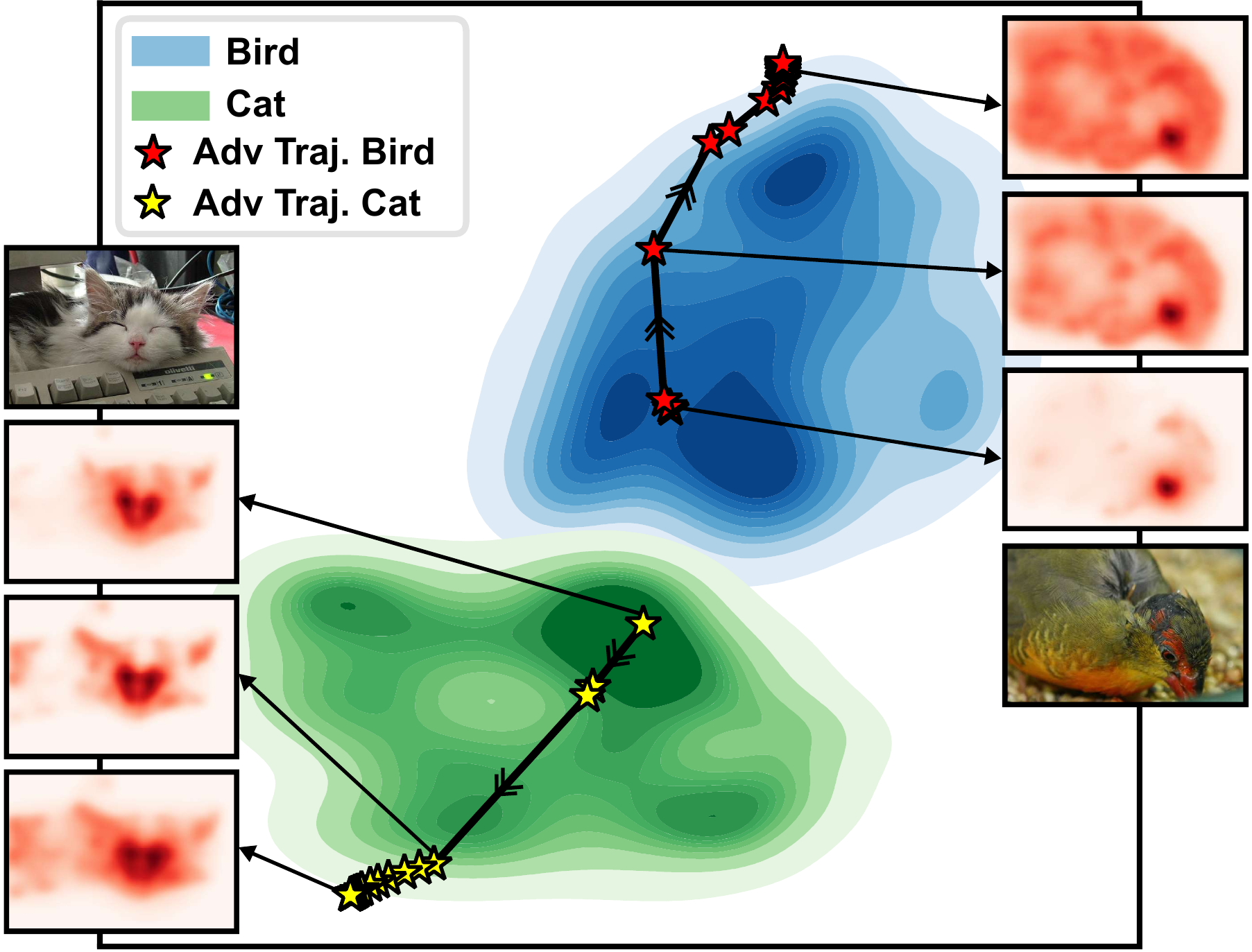}
% \vspace{-1.5em}
\caption{\label{fig_tsne} Feature manifold of images containing the ``bird" (blue) and ``cat" (green), and the trajectory of adversarial climbing for an image from each class. The dimensionality of the feature was reduced by t-SNE~\cite{maaten2008visualizing}.
}
\vspace{-1em}
\end{figure}

\subsection{\rv{Analysis of Results by Class}}\label{classwise_analysis}
\rv{The objects in the images in the \rv{MS COCO 2014} dataset are of various classes with various object sizes. We will now discuss the degree of improvement in the initial seed for each object class.
Fig.~\ref{perclassgain} shows the improvement in mIoU produced by adversarial climbing over the initial seed for each class. The classes are listed in ascending order \rvt{according to} the average size of the target objects in each class (smallest $\rightarrow$ largest).
Adversarial climbing improves mIoU values for the majority of classes, regardless of their average object size.
When considering specific classes, we observed a large drop in the seed quality for the `dining table' class, which is anomalous. We believe that this is due to the ambiguity of the ground truth label of the `dining table'. In \rvt{the MS COCO 2014 dataset}, the `dining table' label includes all the items on the table. The suppression of other classes by the regularization prevents objects such as bowls and bottles on the table from being identified as part of a `dining table', resulting in a localization map that does not entirely match the ground truth.

To take a closer look at how adversarial climbing affects the performance of each class with various object sizes, we \rvt{report} precision, recall, and F1-score values averaged across all classes, the classes corresponding to the 10 smallest objects, and the classes corresponding to the 10 largest objects in Table~\ref{tab_size_pr}.
Our method improves precision, recall, and F1-score of the initial seed, averaged across all classes.
Recall was slightly reduced (-12\%) for the classes corresponding to the 10 smallest objects, but precision increased significantly (67\%), resulting in \rvt{a} largely improved F1-score. 
This indicates that, for small objects, adversarial climbing effectively suppresses unwanted high attribution scores in the background, as can be seen in the `fork' and `handbag' examples in Fig.~\ref{cococamsamples}. 

We believe that there are two causes \rvt{of} these improved results on small objects: 1) During adversarial climbing, the logits associated with classes other than the target are reduced, as described in Section 4.3, and thus patterns which are irrelevant to the target class are effectively suppressed; and 2) since adversarial climbing increases the scores of regions relevant to the target class, the scores of background regions are suppressed due to normalization.

\rvt{Interestingly}, the latter effect is also observed with larger objects. Fig.~\ref{fig_bg_suppress} shows two examples in which adversarial climbing suppresses background regions of images from the CUB-200-2011~\cite{wah2011caltech} dataset. 
Even when the target object is relatively large, we see that the (normalized) attribution score of the background is suppressed as that of the target object is increased.
\jmrv{Fig.~\ref{fig_bg_suppress} shows distributions of attribution scores of each pixel for an AdvCAM and CAM, which are sorted in decreasing order. 
Adversarial climbing widens the gap between the scores of the highly activated regions and those of the remaining regions, which will make the attribution scores of \rvt{the} background reduced after normalization.}

Adversarial climbing improves both precision and recall for large objects, but recall increases by a much larger margin. 
This indicates that adversarial climbing effectively raises the attribution scores of regions of target objects that had not previously been identified.
These observations support our arguments described in Section~\ref{initialseedexp}, namely that our method \rvt{improves} precision for small objects and recall for large objects.

We will now look at how the hyper-parameters interact with the object size. Table~\ref{tab_size_pr} shows the precision, recall, and F1-score values obtained using different values of $T$ and $\tau$. 
Across all classes, neither $T$ nor $\tau$ had a significant influence, which accords with the results presented in Section~\ref{hyperparam}.
Looking at the 10 classes containing the largest target objects, we see a similar picture. However, the 10 classes containing the smallest objects seem to be a little more sensitive to the values of the hyper-parameters, but not sufficiently to be a cause for concern.
}

\subsection{Manifold Visualization}
The trajectory of adversarial climbing can be visualized at the feature level \rv{by} using t-SNE~\cite{maaten2008visualizing} to reduce the dimensionality of each feature.
We assembled a set of images that contain a single object, which is recognized as a `cat' or a `bird' by the classifier. 
We then constructed a set $\mathcal{F}$ containing the features extracted from those images before the final classification layer. 
\rv{Next,} we chose a representative image from each class, and constructed a set $\mathcal{F}~'$ containing the features of those two images together with the features of the 20 versions of each image that resulted from successive manipulations by adversarial climbing.
Fig.~\ref{fig_tsne} shows the features in $\mathcal{F} \cup \mathcal{F}~'$, after dimensional reduction by t-SNE.
We can see that adversarial climbing pushes the features away from the decision boundary that separates the two classes, without causing them to deviate significantly from the feature manifold of their class, even after 20 steps of adversarial climbing.

\section{Conclusions}
We have proposed adversarial climbing, a novel manipulation method to improve localization of a target object by identifying whole regions of the target object from class labels.
An image is perturbed along the pixel gradients of the classifier's output for that image, in a direction that increases the classification score, resulting in \rv{an} attribution map of the manipulated image \rv{that} includes more of the target object.
Because adversarial climbing is a post-hoc analysis of the output of a trained classifier, no modification or re-training of the classifier is required.
Therefore, adversarial climbing can be readily combined with existing methods. We have shown that an AdvCAM, the attribution map generated from adversarial climbing, can indeed be combined with recently developed networks for weakly supervised semantic segmentation and object localization.
The resulting hybrids achieved a new state-of-the-art performance on both weakly and semi-supervised semantic segmentation. 
In addition, the use of adversarial climbing yielded a new state-of-the-art performance on weakly supervised object localization.

\bigskip
\vspace{-0.5em}
\noindent\textbf{Acknowledgements:}
This work was supported by Institute of Information \& communications Technology Planning \& Evaluation (IITP) grant funded by the Korea government (MSIT) [NO.2021-0-01343, Artificial Intelligence Graduate School Program (Seoul National University)], 
% the National Research Foundation of Korea (NRF) grant funded by the Korea government (MSIT) [2018R1A2B3001628], 
AIRS Company in Hyundai Motor and Kia through HMC/KIA-SNU AI Consortium Fund, and the BK21 FOUR program of the Education and Research Program for Future ICT Pioneers, Seoul National University in 2022.
\ifCLASSOPTIONcaptionsoff
  \newpage
\fi



%\bibliographystyle{IEEEtran}


% \end{thebibliography}

{
    \footnotesize
    \bibliographystyle{IEEEtran}
    \bibliography{IEEEabrv,egbib}
}


\begin{IEEEbiography}[{\includegraphics[width=1in,height=1.25in,clip]{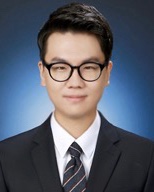}}]{Jungbeom Lee}
received the B.S. degree in electrical and computer engineering from Seoul National University, Seoul, South Korea, in 2017. He is currently pursuing the Ph.D. degree in electrical and computer engineering in Seoul National University, Seoul, South Korea. His research interests include computer vision, deep learning, and label-efficient learning.
\end{IEEEbiography}

\begin{IEEEbiography}[{\includegraphics[width=1in,height=1.25in,clip]{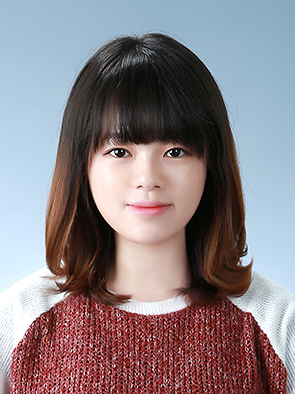}}]{Eunji Kim}
received the B.S. degree in electrical and computer engineering from Seoul National University, Seoul, South Korea, in 2018, where she is currently pursuing the integrated M.S./Ph.D. degree in electrical  and  computer  engineering. Her research interests include artificial intelligence, deep learning, and computer vision.
\end{IEEEbiography}

\begin{IEEEbiography}[{\includegraphics[width=1in,height=1.25in,clip]{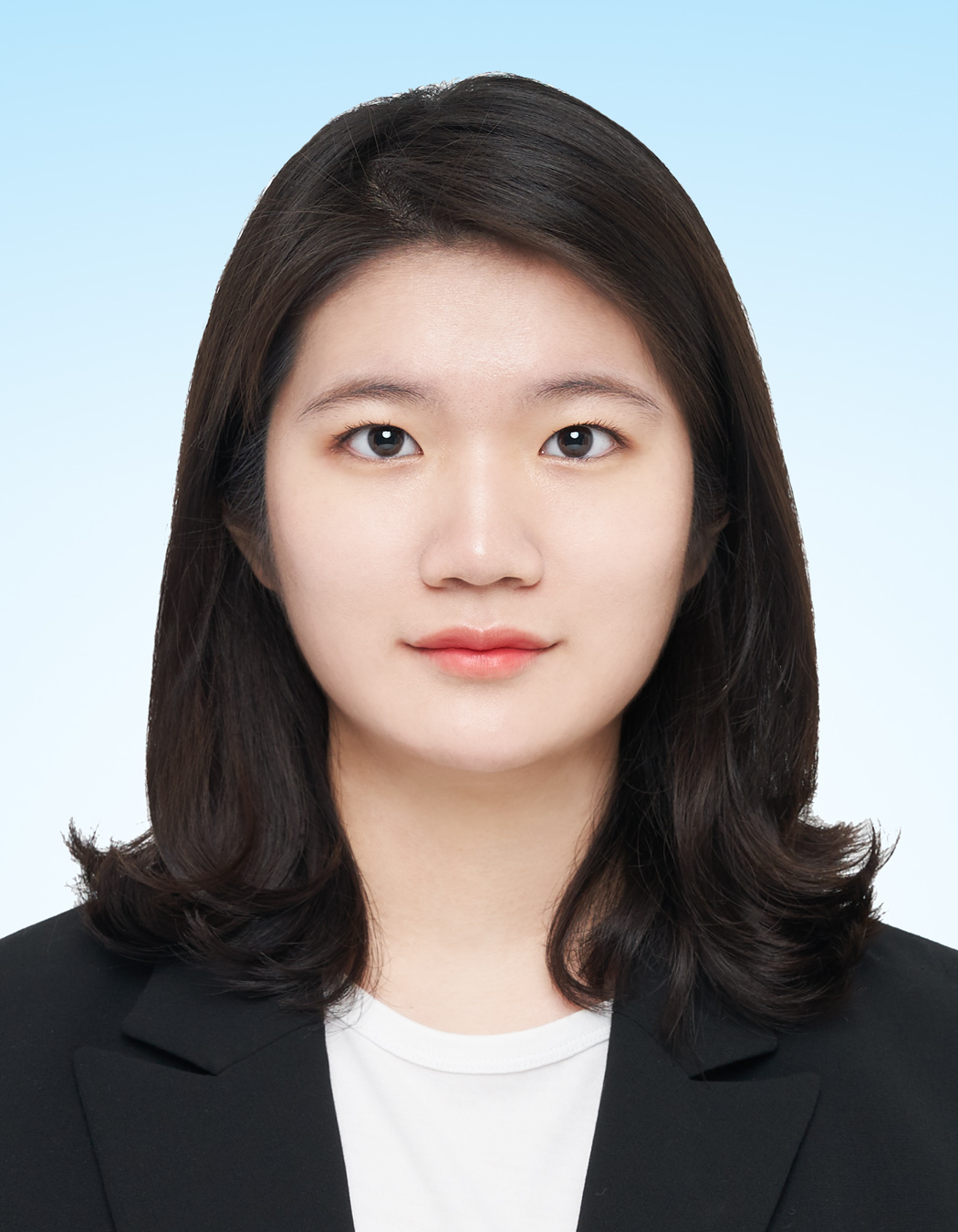}}]{Jisoo Mok}
received the B.S. degree in electrical engineering from California Institute of Technology, CA, USA, in 2018. She is currently pursuing the Integrated M.S./Ph.D. degree in electrical and computer engineering in Seoul National University, Seoul, South Korea. Her research interests include automated machine learning and hardware-aware deep learning.
\end{IEEEbiography}

\begin{IEEEbiography}[{\includegraphics[width=1in,height=1.25in,clip,keepaspectratio]{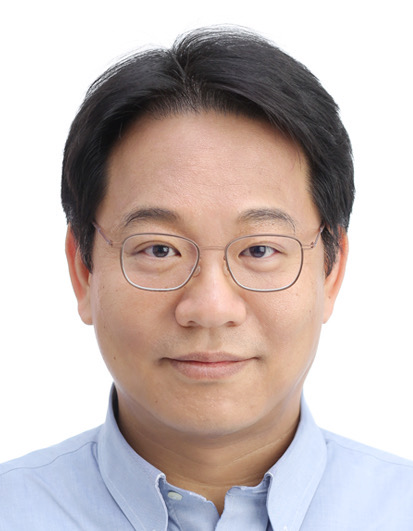}}]{Sungroh Yoon}
(Senior Member, IEEE) received the B.S. degree in electrical engineering from Seoul National University, Seoul, South Korea, in 1996, and the M.S. and Ph.D. degrees in electrical engineering from Stanford University, Stanford, CA, USA, in 2002 and 2006, respectively. He was a Visiting Scholar with the Department of Neurology and Neurological Sciences, Stanford University, from 2016 to 2017. He held research positions at Stanford University and Synopsys, Inc., Mountain View, CA, USA. From 2006 to 2007, he was with Intel Corporation, Santa Clara, CA, USA. He was an Assistant Professor with the School of Electrical Engineering, Korea University, Seoul, from 2007 to 2012. He is currently a Professor with the Department of Electrical and Computer Engineering, Seoul National University. His current research interests include machine learning and artificial intelligence. Dr. Yoon was a recipient of the SNU Education Award, in 2018, the IBM Faculty Award, in 2018, the Korean Government Researcher of the Month Award in 2018, the BRIC Best Research of the Year in 2018, the IMIA Best Paper Award in 2017, the Microsoft Collaborative Research Grant in 2017 and 2020, the SBS Foundation Award in 2016, the IEEE Young IT Engineer Award in 2013, and many other prestigious awards. Since February 2020, he has been serving as the Chairperson for the Presidential Committee on the Fourth Industrial Revolution established by the Korean Government.  
\end{IEEEbiography}


%\vfill


% \clearpage

% \appendices

% % you can choose not to have a title for an appendix
% % if you want by leaving the argument blank
% \section{}

% % use section* for acknowledgment
% \ifCLASSOPTIONcompsoc
%   % The Computer Society usually uses the plural form
%   \section*{Acknowledgments}
% \else
%   % regular IEEE prefers the singular form
% \fi


\end{document}